\documentclass[11pt,fleqn]{report}

\usepackage{amsmath}
\usepackage{graphicx}
\usepackage{float}
\usepackage{xcolor}
\usepackage{listings}
\usepackage{hyperref}

\graphicspath{{figures/}}
\lstdefinestyle{py}{
  language=Python,
  basicstyle=\ttfamily\footnotesize,
  keywordstyle=\color{magenta},
  stringstyle=\color{red},
  emph={abs,range},
  emphstyle=\color{blue},
  showstringspaces=false,
  frame=single,
  rulecolor=\color{lightgray},
  xleftmargin=1em,
  framexleftmargin=1em,
  columns=fullflexible,
  keepspaces=true
}
\lstdefinestyle{shell}{
  basicstyle=\ttfamily\footnotesize,
  columns=fullflexible,
  keepspaces=true,
  xleftmargin=2em
}

\begin{document}
\hypersetup{pageanchor=false}

\begin{titlepage}
  \centering
  {\Large\bfseries Adaptive Distance Detection and\\
  Adaptive Instance Segmentation\par}
  \vfill
  {\Large\bfseries Kai Yi Ji\\
  2178893\\
  Bsc Computer Science\par}
  \vspace{1cm}
  {\Large\bfseries Supervisor: Ahmad Ibrahim\\
  Inspector: Leonardo Stella\par}
  \vfill
\end{titlepage}

\thispagestyle{empty}
\begin{abstract}
This paper is about object detection model and semantic segmentation model
in an FPS game Counter Strike: Condition Zero to detect enemies. Semantic
segmentation is pixel-wise detection algorithm. We customized the architecture
for semantic segmentation. We used YOLOv5s to train object detection
model. After training, the inference time of semantic segmentation was compared
to object detection. It was observed that object detection was slightly
faster and semantic segmentation is slower but better at detecting the exact
shape of the object. However, we found out that this slight difference of
inference can grow exponentially if the both models get more complex. To
solve this, we propose a novel approach called Adaptive distance detection
which combines both algorithms. We will also introduce Adaptive instance
segmentation to combine semantic segmentation and panoptic segmentation.
\end{abstract}
\clearpage

\hypersetup{pageanchor=true}
\pagenumbering{arabic}
\tableofcontents

\chapter{Introduction}

There has been a rise of research papers about AI. One such field of AI is
Computer Vision. Its uses include transportation, healthcare, manufacturing,
agriculture, and many others. In this project, we try to demonstrate the
potential use of detection algorithms from computer vision in a game. We
will compare inference time of modern detection models. We will also attempt
to create a custom model which uses the combination of models with
different inference times in an FPS shooting game Counter Strike: Condition
Zero. Later in the paper we introduce the novel architecture to combine
detection algorithms.

\section{Project Aims}

This project firstly aims to compare the latency of different algorithms for
detection. This includes object detection (YOLOv5s), semantic segmentation
(custom). We will observe the inference time of these algorithms. In
addition, measures such as accuracy will be also presented. Our focus will
mainly be the comparison of semantic segmentation and object detection.
Then the project introduces our new architecture which combines semantic
segmentation and object detection, which is called adaptive distance detection
to balance between accuracy and latency. We are also going to combine
semantic segmentation with panoptic segmentation named adaptive instance
segmentation.

\section{Background}

In this section we will provide enough background concepts to provide context
to our research.

\subsection{Enemy Detection in CS: Condition Zero}

Counter Strike was a building block for modern FPS shooting games. Its
game mechanics is very similar to modern despite it being a very old game.
The player is assigned to a team whether as a terrorist or counter terrorist.
The objective of terrorists is to plant the bomb and make sure it detonates.
The objective of counter terrorists is to defuse the bomb. It is a very complex
game with different tactics being involved such as managing economy for the
team, positioning your player, aiming enemies. Aiming has always been the
core techniques of this game. As a player, you must detect an enemy, try
to aim at the enemy (whether it is head or body) and shoot at him. A
good player will need to have a good reaction to have advantage against
other players. But before that, you first need to detect enemy players to be
aim. This process is often simple for us humans, but from perspective of
machine, it goes through multiple stages to achieve that. In this project, we
will create an AI that is able to detect enemy players. In computer vision, we
have different algorithms (object detection, segmentation) to detect specific
objects. Each has its own benefits. We will use both techniques to detect
enemies. Moreover, we would also need to be aware of time of inference
of those algorithms (an analogy to human reaction) to get advantage as an
artificial intelligence.

\subsection{Convolutional neural network}

A typical neural network has an input layer, hidden layers as well as an
output layer. We pass the input data to the input layer then the network
will pass that data through series of computations using hidden layers to
output layer, where the result is predicted. After prediction we compute the
loss function between predicted value and ground truth value. After that,
we backpropagate our loss from the output layer to input layer to update all
the necessary parameters with the help of gradients. Convolutional neural
network (CNN) is one type of neural network that works with images. We
typically represent image as H x W x 3 matrix (RGB image) or as HxWx1
matrix (Grayscale image). We then use kernels of size F x F to convolve
with the image. Convolution operation extracts features from the image
such as edge shape, color and passes them to the next layer. We repeat this
process until we get to output. Depending on the task we may use different
variations of CNN. Classification, object detection, semantic segmentation
all use CNN to classify, detect and segment images.

\subsection{Activation Function}

An activation function is used to decide whether a neuron should be activated
in a network. Think of it as an analogy to our brain. When we receive specific
certain information from vision, our brain tries activating certain neurons to
pass the information. The same principle also applies to neural network. We
have different variations of activation function; each has its own advantages.
For example, Sigmoid activation \cite{pratiwi2020sigmoid} and ReLU \cite{agarap2019relu} activation were used in
this project. The benefit of sigmoid \cite{pratiwi2020sigmoid} function is that activated value
does not blow up but in exchange it increases computing time. On the
other hand, ReLU \cite{agarap2019relu} is shown to have better efficiency and leads to faster
convergence but also eliminates vanishing gradient. Vanishing gradient is a
problem during backpropagation when the gradient is very small making it
difficult to update weights.

\subsection{Pooling Layer}

Pooling layer is often added to convolutional neural network to reduce the
dimensions from previous so that computation is more efficient. For instance
we have Max Pooling \cite{wu2015maxpool} layer that reduces the dimensions by choosing the
maximum value of filters. The data is downsampled and can be computed
more efficiently.

\subsection{Optimizers in Neural Network}

During learning of neural network, the weights are updated from loss function.
However, we can choose how weights are updated by choosing an optimizer
such as Nesterov Gradient Descent, Adagrad \cite{duchi2011adagrad} , RMSProp. Nesterov
gradient descent uses its momentum to arrive at global optimum faster. Adagrad
is very well suited for sparse data with decaying learning rate, however
after a number of iterations, it slows down the update. RMSProp is introduced
to prevent the slowdown of Adagrad by. We decided to use Adam
\cite{kingma2017adam} optimizer, which combines both RMSProp and Nesterov optimizer and
is commonly the best choice to train the model. Further observations on
optimizers is presented in \cite{ruder2017overview}.

\subsection{Loss Function}

A loss function is a method of how the model calculates the error between
predicted data and actual data. Based on different learning tasks, we may
use different loss functions. Namely, binary cross entropy \cite{ruby2020bce} is the standard
loss function in image detection. It is the probabilistic method that calculates
the distance between the predicted value and truth value. An overview of
loss functions is presented in \cite{patel_loss}.

\subsection{Regularization}

Regularization is a method to generalize the model to predict unseen data.
This prevents model from overfitting, which means model being too complex
and unable to predict unseen data.

\section{Related Works}

Numerous papers have been published on developing artificial intelligence
for playing games, such as \cite{openai2019dota}, which is about an large scale AI trained
with reinforcement learning algorithm by OpenAI to play against champions
in DOTA2. The artificial intelligence won by 2-0 against human players,
showing its incredible capabilities. Similar concept was then applied to other
games, such as \cite{silver2016go}, \cite{vinyals2019starcraft} for Go and StarCraft. As for FPS games we have
papers \cite{mcpartland2008fps, liang2018fps}, which use reinforcement learning to create a multi-purpose
bots. Other papers, such as \cite{pearce2021csgo, witschel2020aimlow} which are trying to clone human behaviour
to make the bots more believable as a human. \cite{kanervisto2022gan} uses GAN, which
consists discriminator and generator to train the bot. Our project is going
to be more similar to \cite{mahendran2016doom, jung2021gamescenes, venkatesh2018tracking}, which uses detection algorithms consisting
CNN to track objects.

Image classification is a fundamental field that assigns a label to a whole
image. There are exists number of papers such as \cite{simonyan2015vgg, lecun1998gradient, krizhevsky2012alexnet, szegedy2014inception, wang2019cspnet}. \cite{simonyan2015vgg}
evaluated convolutional networks with increasing depths with 3x3 filters. \cite{krizhevsky2012alexnet}
has achieved second best performance in an ImageNet LSVRC-2010 contest.
\cite{szegedy2014inception} created a very deep convolutional network with 22 layers deep while
keeping computational resources constant. \cite{wang2019cspnet} was proposed to speed up
the inference. \cite{he2015resnet} simplifies training on deep networks. It presented a residual
block that has a shortcut connections to further layers. \cite{howard2017mobilenets} presented
class of MobileNets that is based on depth-wise separable convolutions. \cite{tan2020efficientnet}
studied that balancing network width, depth and resolution will lead to better
performance. It then created a baseline network that presented a family
of network, named EfficientNets. \cite{he2015resnet, tan2020efficientnet, howard2017mobilenets} are widely used as backbone
for scaling up further networks.

Object detection is another fundamental task for deep learning. The
caveat is that the output of neural network gives x1, x2 and y1, y2 coordinates
to locate object so that the bounding box can be created. Papers such
as \cite{girshick2014rcnn, girshick2015fast, dai2016rfcn, redmon2016yolo} proposed the models that use this algorithm. \cite{girshick2014rcnn} takes
regions of interests as input and then outputs probabilities per class. \cite{girshick2015fast}
is the improved version of \cite{girshick2014rcnn}, possessing an additional CNN to speed up
performance. \cite{dai2016rfcn} proposed position-sensitive score maps to solve translation
invariance in classification and variance in object detection. This achieved
sharing of computation on the entire image, resulting faster than \cite{girshick2015fast}. \cite{redmon2016yolo}
is the model we are going to use in this paper. \cite{redmon2016yolo} solves object detection as
a regression problem, using single neural network to predict bounding boxes
and class probabilities within just one evaluation, achieving extremely fast
performance.

Semantic segmentation is a way of detecting objects by assigning each
of pixel of the object to its category.\cite{ronneberger2015unet, lin2017fpn, long2015fcn, takikawa2019gscnn} and \cite{wu2019fastfcn} are often used
as an architecture for Semantic segmentation. These architectures often
use down-sampling and up-sampling method to process image data.\cite{long2015fcn} is
a network that takes arbitrary input to segment the output. However, this
network does not use downsampling to reduce the dimensions of the image.
This can, in fact, have huge performance cost on the system. \cite{ronneberger2015unet} contains
downsampling to capture context and upsampling to enable precise localization.
Since \cite{ronneberger2015unet} has donwsampling, it effectively reduces the dimensions
of the image to get the context, therefore reducing computation. \cite{lin2017fpn} has
a pyramid hierarchy architecture to scale feature mappings top-down. \cite{takikawa2019gscnn}
added gates to its architecture to remove noise, effectively having sharper
predictions. The paper introduced gated convolutional layers to the architecture.
It has regular stream, shape stream that consists gated convolutional
layers. Then the fusion module combines information from regular stream
and shape stream. LinkNet \cite{chaurasia2017linknet} is an efficient semantic segmentation architecture
that consists of encoder and decoder blocks to segment the image.
\cite{xie2021segformer} proposed a simple segmentation architecture that unified Transformers
with multilayer perception decoders. This led to efficient performance on
segmentation on Transformers.

Panoptic segmentation \cite{kirillov2019panoptic} uses the same concept as semantic segmentation
except it also differentiates instances of detected objects. \cite{cheng2021maskformer} showed
excellent results using mask classification model for semantic and panoptic
segmentation tasks.

Dataset has been a huge part for training a neural network. There have
been huge efforts put into data collection to cover wide range images. Both
\cite{zhou2017ade20k} and \cite{zhou2018ade20k} are the papers that proposed ADE20K dataset. \cite{zhou2017ade20k} uses a novel
Cascade Segmentation Module to parse scenes. Further analysis and benchmarking
were conducted on the research \cite{zhou2018ade20k}. \cite{lin2014coco} is yet another dataset that
is gathered from everyday scenes.

Libraries are frequently used to develop deep learning frameworks. Pytorch
\cite{paszke2019pytorch} and Tensorflow \cite{abadi2015tensorflow} are the most used libraries in recent years.
These libraries simplifies the workflow by focusing on the usability of the
libraries. This makes both libraries very intuitive to use.

This paper takes inference time into consideration when using object detection,
semantic segmentation and panoptic segmentation. Each algorithm
has its own benefits and drawbacks. Therefore, we present an architecture
that balances detection algorithms using a classifier.

\chapter{Experiment Study}

\section{Metrics}

\subsection{Machine Settings}

We are going to measure the inference time of all models using Processor Intel(R)
Core(TM) i7-1065G7 CPU @ 1.30GHz, 1498 Mhz, 4 Core(s), 8 Logical
Processor(s). However, training all models with CPU can be excessive. For
this reason, we will also use Google Colab \cite{bisong2019colab}. Google Colab \cite{bisong2019colab} is a service
from Google to allow us to write python code for Machine learning tasks
and provides free GPU/CPU usage. However, there is also a limit to GPU
resources. Therefore, we need to use GPU sparingly. After training our
model we will export weights and use it to observe inference time on local
computer. We change the configuration to allow CPU inference in the models.
Note that we intentionally chose CS: Condition Zero instead of CSGO
for measurement because our machine was not able to run smoothly CSGO
with Intel Iris Plus.

\subsection{Inference Time}

Our main metric will be the inference time (latency), which is the amount of
time it takes for the machine to output prediction. This is crucial because an
algorithm must perform fast to eliminate the enemy first. Additionally, we
will provide the accuracy parameter for semantic segmentation and object
detection models. In addition to measuring the inference time of models run
on our local computer, we will introduce common metrics for the accuracy
of the model.

\subsection{Accuracy}

Accuracy is a common metric to measure the performance of machine learning
models. It represents the fraction of correct predictions to total number
of predictions. Its formula is written in Equation \ref{eq:accuracy}. TP means true positive,
which means the number of times the machine made correct positive
prediction. In this formula, TP stands for true positives, which is the number
of times the machine made correct positive predictions. TN (true negatives)
indicates the number of correct negative predictions made by the machine.
FP (false positives) and FN (false negatives) refer to the number of incorrect
positive and negative predictions made by the machine.
\begin{equation}
  Accuracy = \frac{TP + TN}{TP + TN + FP + FN}
  \label{eq:accuracy}
\end{equation}

\subsection{Precision}

Precision measures the quality of positive predictions model made. It is the
number of TP divided by TP + FP, as presented in Equation \ref{eq:precision}.
\begin{equation}
  Precision = \frac{TP}{TP + FP}
  \label{eq:precision}
\end{equation}

\subsection{Recall}

Recall, like precision, represents the proportions of actual positives identified
correctly. It is the number of TN divided by (TN + FN). This is illustrated
in Equation \ref{eq:recall}.
\begin{equation}
  Recall = \frac{TP}{TP + FN}
  \label{eq:recall}
\end{equation}

\subsection{F1-score}

F1-score (Dice coefficient) is the harmonic mean of precision and recall, it
is (2 * P * R) / P + R where P is precision and R is recall. Equation \ref{eq:f1pr}
illustrates F1-score in terms of precision and recall. Equation \ref{eq:f1tp} presents
F1-score in terms of TP, FP, FN.
\begin{equation}
  F1Score = \frac{2 \times precision \times recall}{precision + recall}
  \label{eq:f1pr}
\end{equation}
\begin{equation}
  F1Score = \frac{TP}{TP + \frac{1}{2}(FP + FN)}
  \label{eq:f1tp}
\end{equation}

\subsection{Intersection over Union (IoU)}

Intersection over Union is another important metric to measure accuracy of
a model. It is used to describe the overlap of predicted area and truth area.
We can represent IoU as shown in Equation \ref{eq:iou}.
\begin{equation}
  IoU = \frac{TP}{TP + FN + FN}
  \label{eq:iou}
\end{equation}

\section{YOLO : You Only Look Once}

\subsection{Experiment}

Our first experiment begins with a famous object detection algorithm YOLO
\cite{redmon2016yolo}. \cite{redmon2016yolo} states ``A single neural network predicts bounding boxes and class
probabilities directly from full images in one evaluation.'' We decided to use
this algorithm because of its latency performance and accuracy. Its light
version Fast YOLO was claimed to have 155 FPS which is an impressive
performance, better than R-CNN \cite{girshick2014rcnn}. Though it is prone to localization
errors according to the paper. We are going to be using YOLOv5s from
Ultralytics \cite{jocher2020yolov5}. Firstly, we need to have a dataset that we can train the
model. We found out that there exists a dataset in Roboflow \cite{roboflow} already
having images for counter strike \cite{cs16dataset} . It has 173 training images, 25 test
images and 50 validation images. Interestingly, the codebase automatically
generates figures after training the dataset. In addition, we measure the
inference time of model. We trained the model with our dataset which gave
us the following results with batch size of 16 and 100 epochs.

\subsection{Experimental Results of YOLOv5s}

\begin{figure}[H]
  \centering
  \includegraphics[width=13cm]{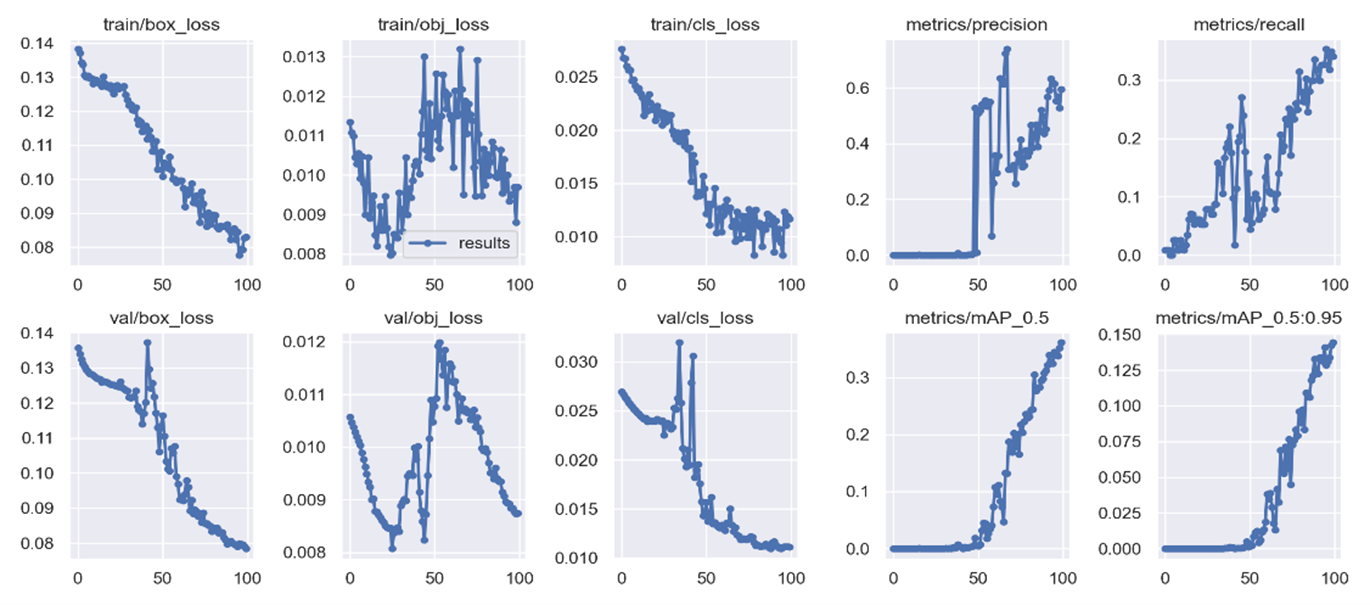}
  \caption{Loss and accuracy}
  \label{fig:2.1}
\end{figure}

\begin{figure}[H]
  \centering
  \includegraphics[width=10cm]{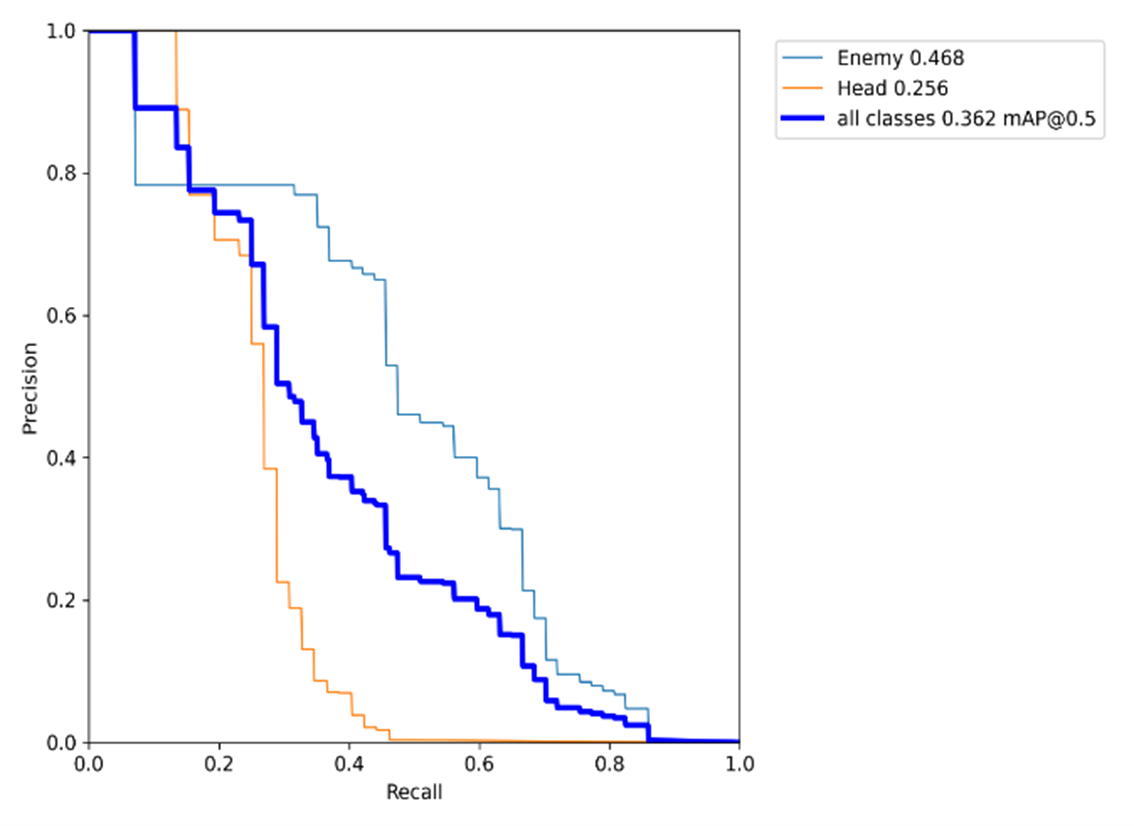}
  \caption{Precision to recall}
  \label{fig:2.2}
\end{figure}

\begin{figure}[H]
  \centering
  \includegraphics[width=10cm]{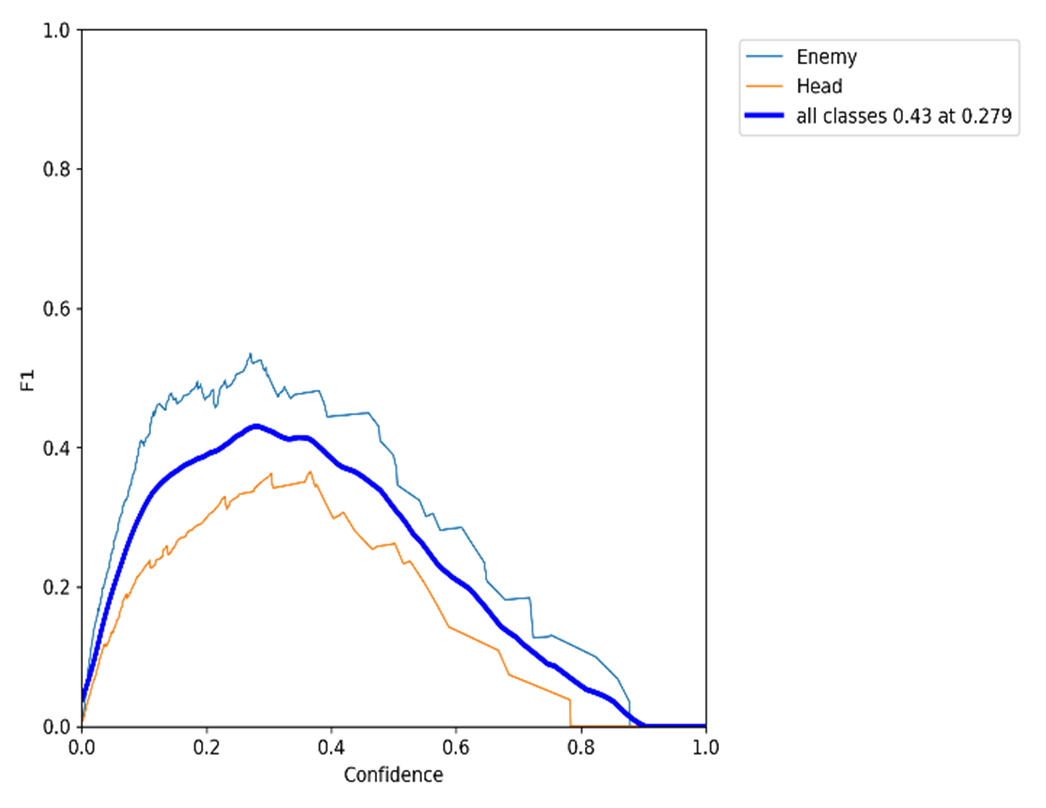}
  \caption{F1 score}
  \label{fig:2.3}
\end{figure}

\begin{figure}[H]
  \centering
  \includegraphics[width=10cm]{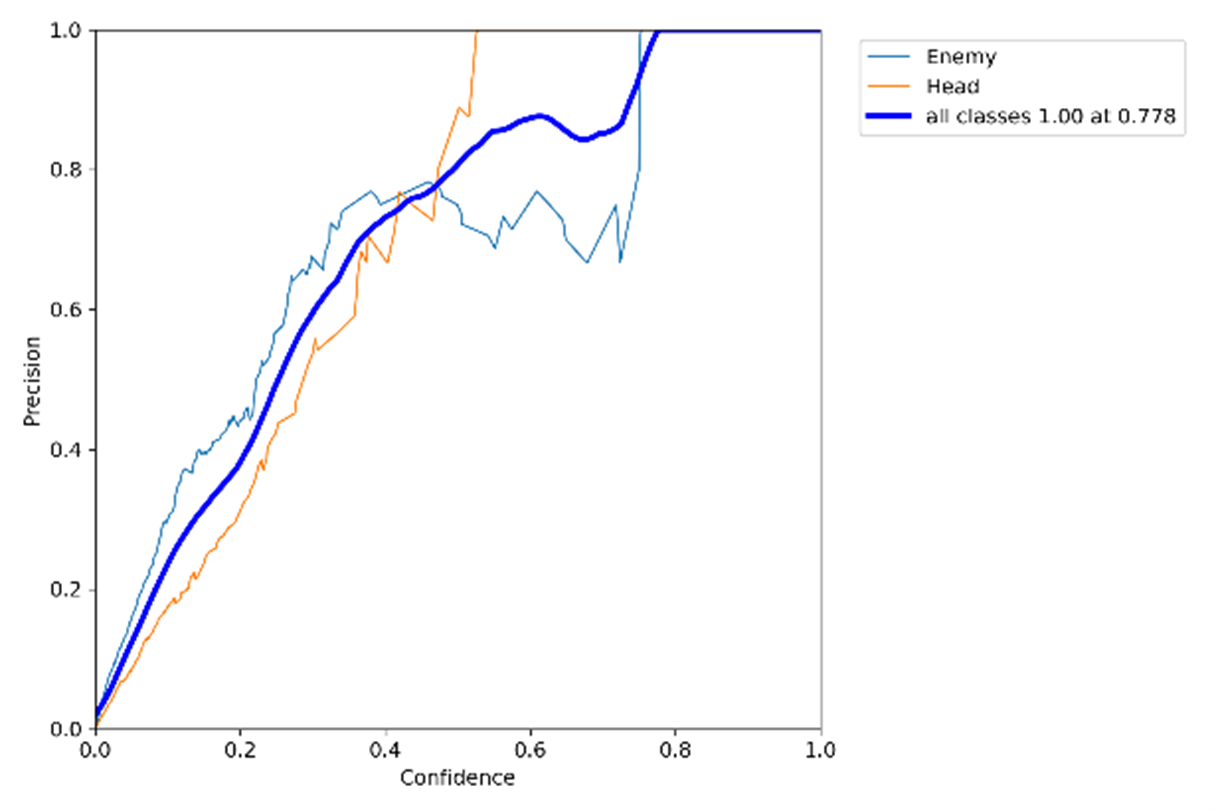}
  \caption{Precision to confidence}
  \label{fig:2.4}
\end{figure}

\begin{figure}[H]
  \centering
  \includegraphics[width=10cm]{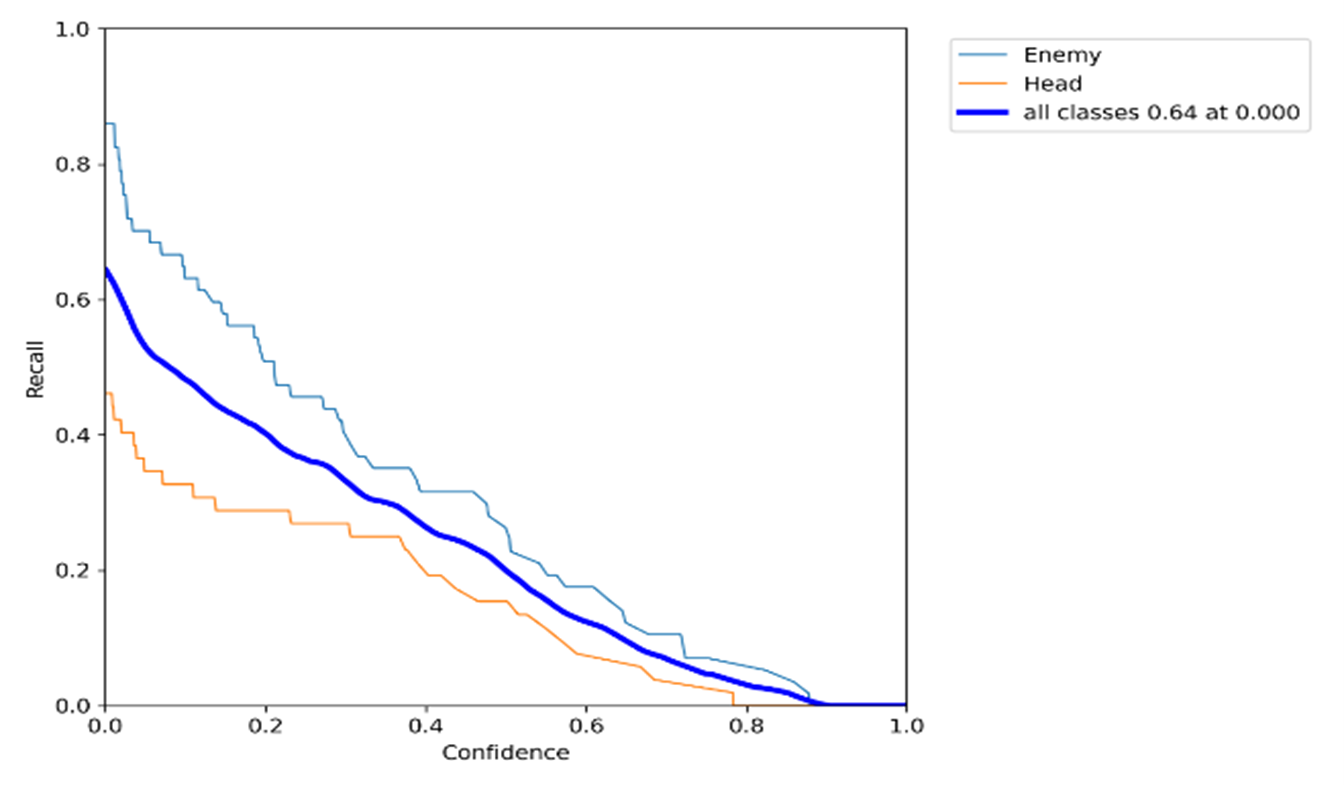}
  \caption{Recall to confidence}
  \label{fig:2.5}
\end{figure}

Figure \ref{fig:2.1} shows the loss of our model. We see our train/box\_loss heading
towards 0.08. We also focus on metric such as precision, recall and mAP
since they provide accuracy for our model. We observe that precision is
just right below 0.60 and recall is above 0.30. We have Figure \ref{fig:2.2} to show
tradeoff between precision and recall. Generally, if area under the curve is
big then we know that it has both high precision and high recall, which
means the model is very accurate. Figure \ref{fig:2.3} illustrates F1-score along with
confidence. The confidence serves as a parameter to which how certain is
the model with predicted bounding boxes. If the confidence is below certain
threshold then the model discards the bounding box. We observe that F1
peaks at confidence 0.279 with F1-score 0.43. This means the model is most
accurate when it has 0.279 confidence. Subsequently, we have Figures \ref{fig:2.4}
and \ref{fig:2.5} to describe recall and precision to confidence. We observe that recall
is at its best when confidence is 0 with 0.64. This is because we confidence
is low we do not discard any area and therefore the number of FN will be
very low. The opposite is precision to confidence curve , which is at its worst
when confidence is 0 but best at 0.778. This is because when confidence is 0
we have high FP, hence not performing well.

\section{Semantic Segmentation}

\subsection{Custom Dataset}

As for semantic segmentation, we did not find any existing dataset that has
segmented images of CS: Condition Zero. Therefore, a custom dataset is
created for segmentation training. We used Label studio \cite{tkachenko2020labelstudio} as a software
to annotate all of segmentation images that were captured. Label studio \cite{tkachenko2020labelstudio}
is a well-known library to label data manually and is not limited to images.
You can use different methods to label data such as polygon labeling or
brush labeling. In this dataset, brush labeling was used to annotate the
data. After that, we exported our data as png format. We then split our
dataset to training, validation, testing sets with 20, 5, 8 images respectively.

\subsection{Experiment}

We will be importing code from \cite{iakubovskii2019smp} to customize our model . It has prepared
common segmentation architectures for us, such as Unet \cite{ronneberger2015unet} and FPN \cite{lin2017fpn}
as well as backbones, such as \cite{tan2020efficientnet} and \cite{howard2017mobilenets}. We need to change parts of the
code to solve our particular task. We also need to choose our loss function.
We consider two loss functions, particularly Dice loss and Jaccard loss \cite{bertels2019dice}.
Dice loss is widely used to handle class imbalance problems. It is derived
from Dice coefficient, which is another term for F1-score. Jaccard loss is derived
from Jaccard index, which is another term for Intersection over Union.
We conduct an experiment to compare Jaccard loss to Dice loss. To do that,
we will use backbone efficientnetb3 \cite{tan2020efficientnet}. This backbone, compared to MobileNets
\cite{howard2017mobilenets} and ResNet \cite{he2015resnet} , achieves both high efficiency and accuracy.
After comparing loss functions, we then procede to compare different architectures,
such as \cite{ronneberger2015unet, lin2017fpn, chaurasia2017linknet}. Since we are focusing on binary segmentation
task, we will use sigmoid \cite{pratiwi2020sigmoid} activation function. We train our model with
batch size of 8, learning rate 0.0008 and 80 epochs. For measuring accuracy
in test dataset, we use IoU (Intersection over Union) and F1-score.

\subsection{Loss Results and Comparison}

\begin{figure}[H]
  \centering
  \includegraphics[width=12cm]{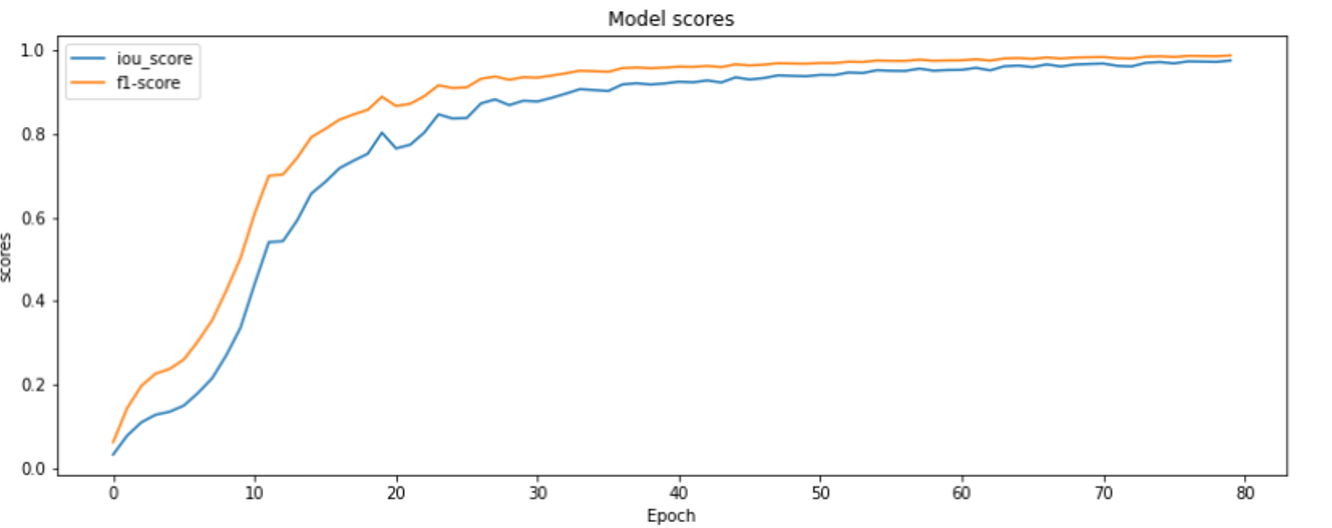}
  \caption{Accuracy of Jaccard loss}
  \label{fig:2.6}
\end{figure}

\begin{figure}[H]
  \centering
  \includegraphics[width=11.7cm]{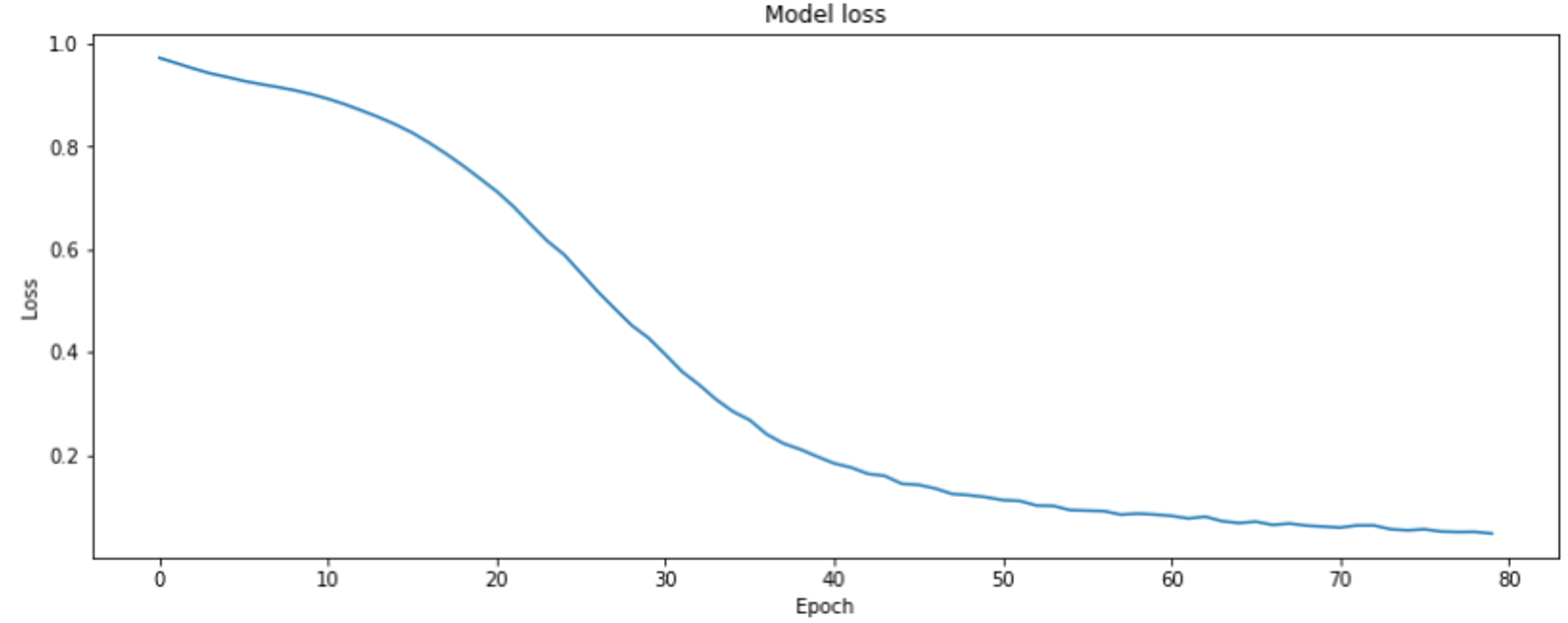}
  \caption{Jaccard loss}
  \label{fig:2.7}
\end{figure}

\begin{figure}[H]
  \centering
  \includegraphics[width=12cm]{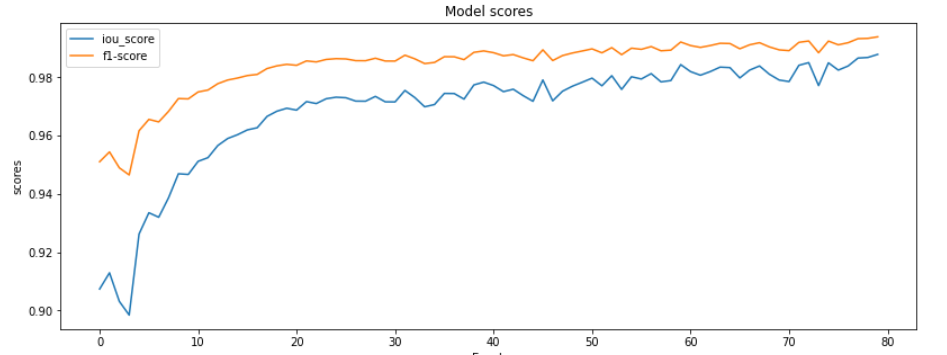}
  \caption{Accuracy of Dice loss}
  \label{fig:2.8}
\end{figure}

\begin{figure}[H]
  \centering
  \includegraphics[width=12cm]{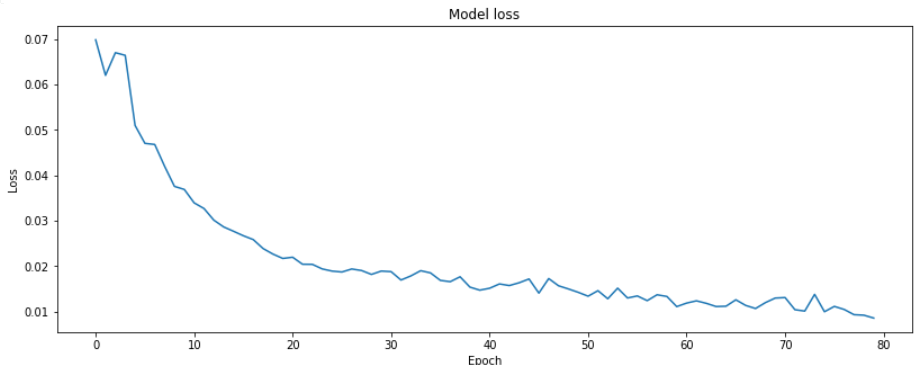}
  \caption{Dice loss}
  \label{fig:2.9}
\end{figure}

During experiment, we observed that both Dice loss functions and Jaccard
loss achieve high accuracy when converging. Figure \ref{fig:2.6} shows the accuracy
of Jaccard loss function. Compared to Figure \ref{fig:2.8}, which is the accuracy
of Dice loss, we observe that Jaccard was more stable at converging than
Dice loss. Figures \ref{fig:2.7} and \ref{fig:2.9} show the Jaccard and Dice loss respectively,
and we can verify that Dice loss was more unstable than Jaccard loss, when
minimizing. During testing, both loss function had high accuracy. However,
Jaccard loss had slightly better performance than Dice loss, as shown in
Table \ref{tab:2.1}. Because of this, we decided to use Jaccard loss.

\begin{table}[H]
  \centering
  \caption{Test accuracy for loss functions}
  \label{tab:2.1}
  \begin{tabular}{|p{3cm}||p{3cm}|p{3cm}|}
    \hline
    \multicolumn{3}{|c|}{Test accuracy} \\ \hline
    Loss function & Mean IoU & Mean F1-score \\ \hline
    Jaccard & 0.70 & 0.82 \\
    Dice    & 0.64 & 0.78 \\ \hline
  \end{tabular}
\end{table}

\subsection{Architecture Results and Comparison}

\begin{figure}[H]
  \centering
  \includegraphics[width=12cm]{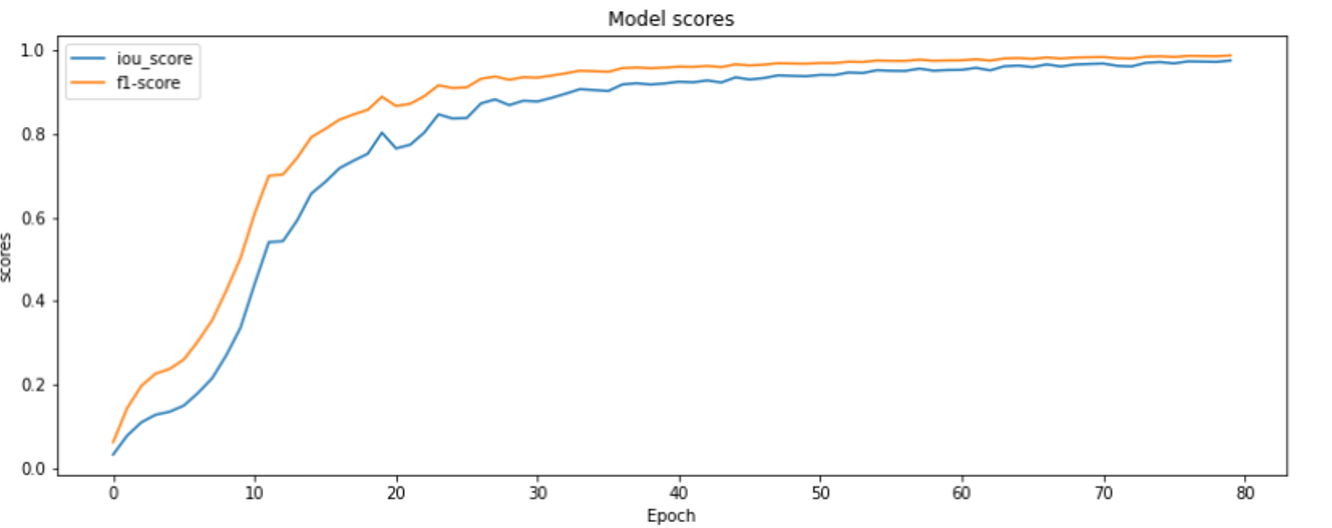}
  \caption{Accuracy of U-Net}
  \label{fig:2.10}
\end{figure}

\begin{figure}[H]
  \centering
  \includegraphics[width=11.7cm]{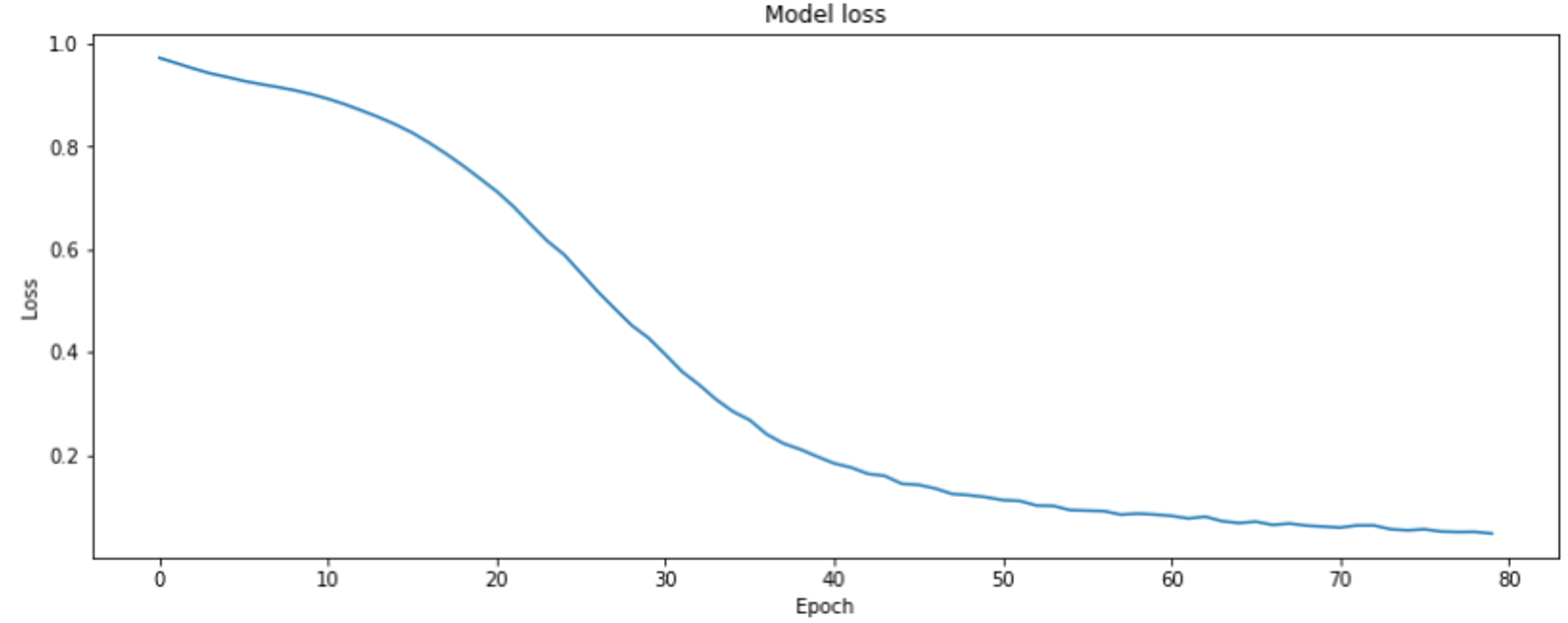}
  \caption{U-Net loss}
  \label{fig:2.11}
\end{figure}

\begin{figure}[H]
  \centering
  \includegraphics[width=12.3cm]{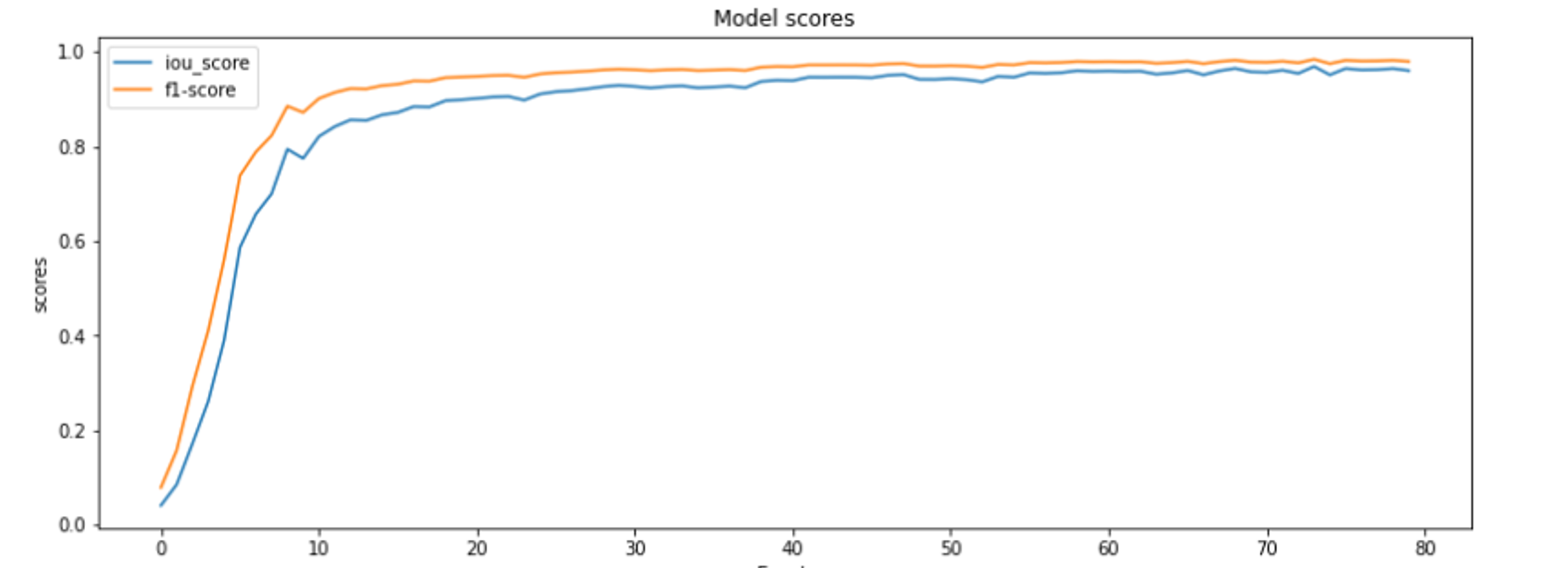}
  \caption{Accuracy of FPN}
  \label{fig:2.12}
\end{figure}

\begin{figure}[H]
  \centering
  \includegraphics[width=12cm]{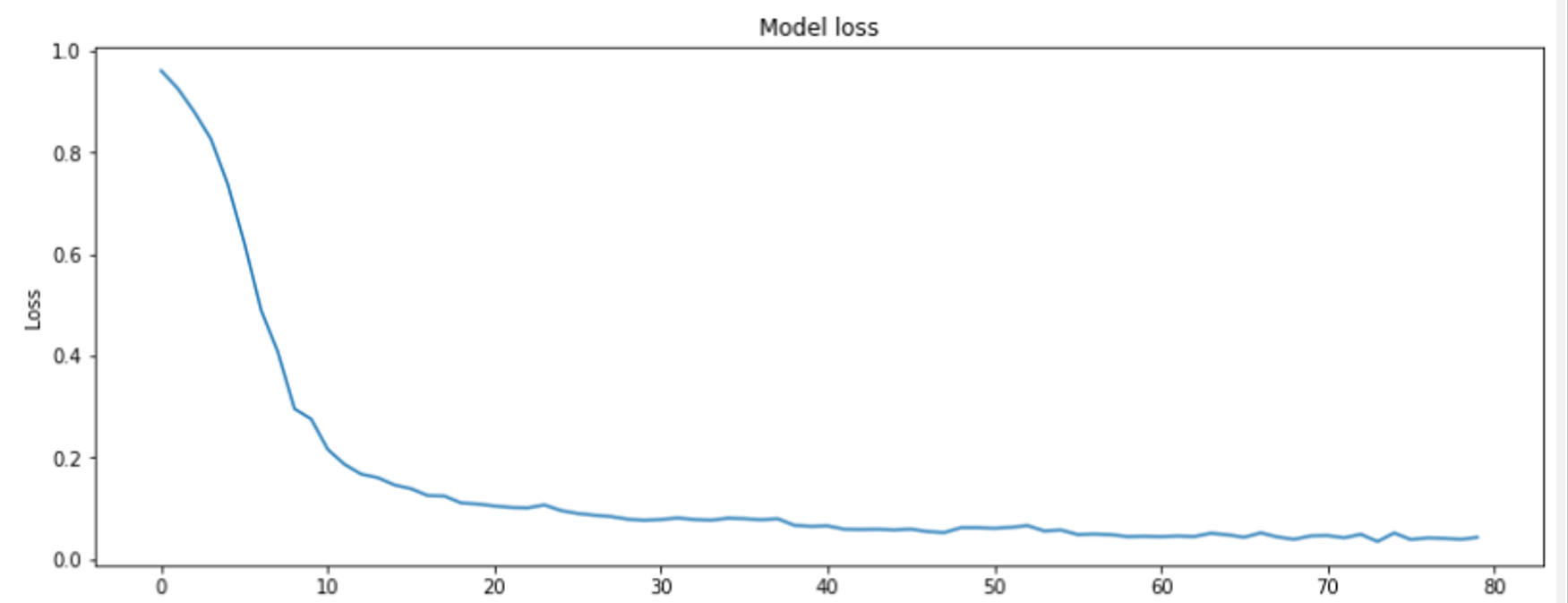}
  \caption{FPN loss}
  \label{fig:2.13}
\end{figure}

\begin{figure}[H]
  \centering
  \includegraphics[width=12.6cm]{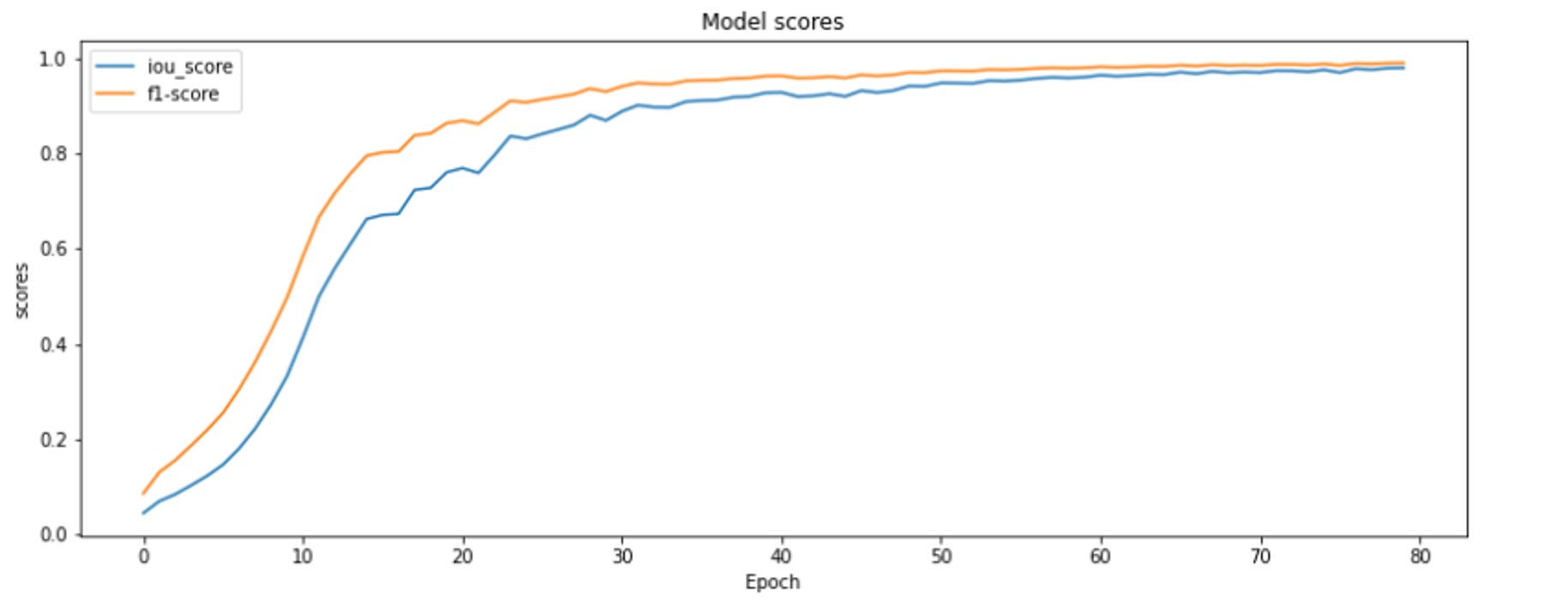}
  \caption{Accuracy of LinkNet}
  \label{fig:2.14}
\end{figure}

\begin{figure}[H]
  \centering
  \includegraphics[width=12cm]{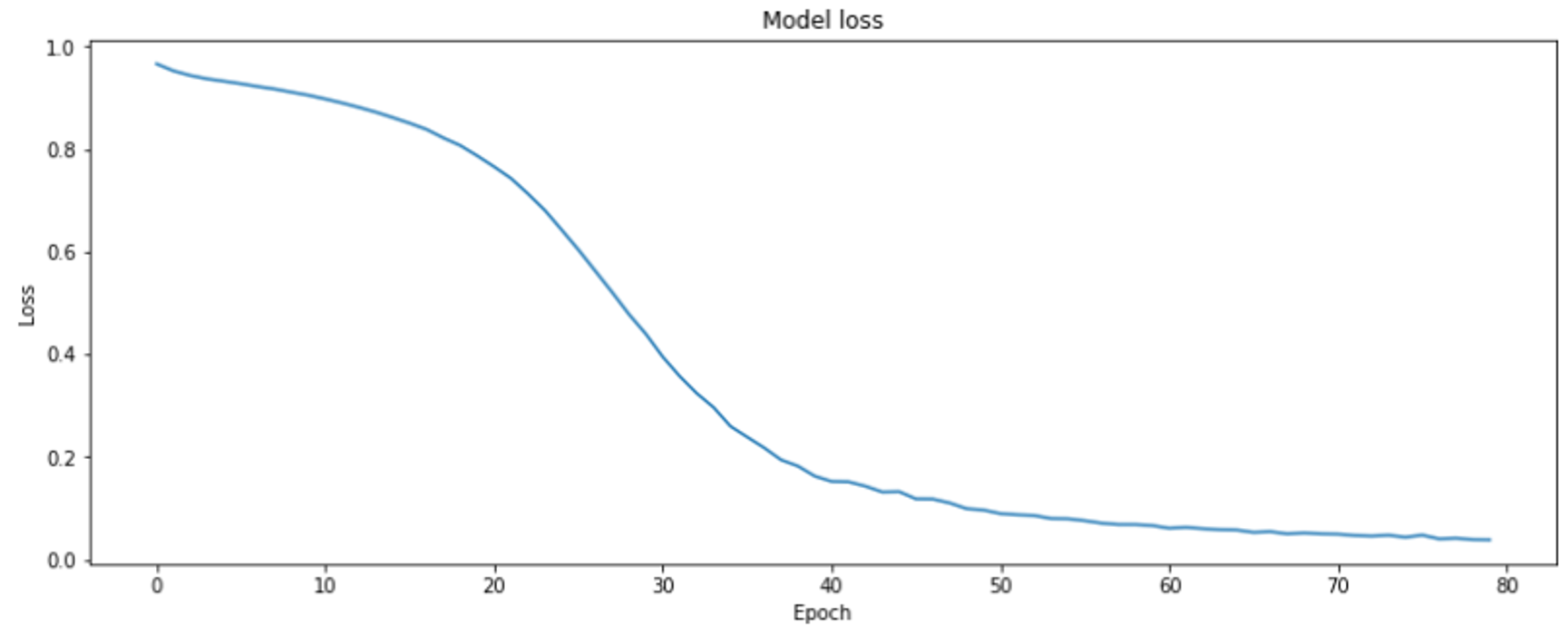}
  \caption{LinkNet loss}
  \label{fig:2.15}
\end{figure}

\begin{figure}[H]
  \includegraphics[width=15.2cm]{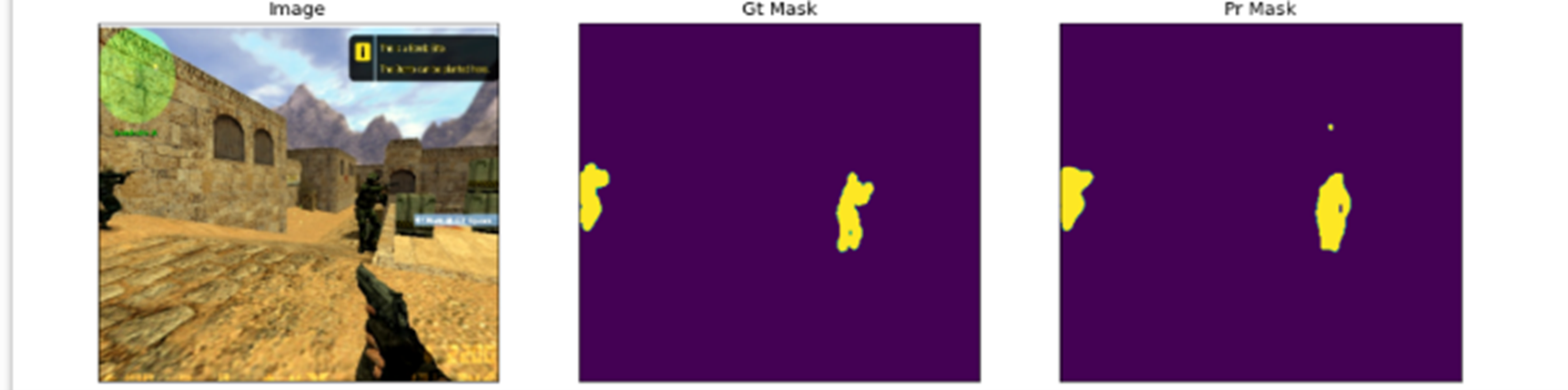}
  \caption{Segmentation example of U-Net architecture}
  \label{fig:2.16}
\end{figure}

The accuracy and loss of U-Net \cite{ronneberger2015unet} architecture are illustrated in Figures
\ref{fig:2.10} and \ref{fig:2.11}. We note that Figures \ref{fig:2.10} and \ref{fig:2.11} are indentical to previous
Figures \ref{fig:2.6} and \ref{fig:2.7} when comparing loss functions since we used U-Net
architecture to compare loss functions. Figures \ref{fig:2.12} and \ref{fig:2.13} show the accuracy
and loss of FPN \cite{lin2017fpn} architecture. We noticed that during training,
FPN model was fastest to converge of all architectures. We see that U-Net
converged slower than FPN. LinkNet \cite{chaurasia2017linknet} was slowest at converging, as depicted
in Figures \ref{fig:2.14} and \ref{fig:2.15}. Table \ref{tab:2.2} shows IoU score and F1-score of
architectures. FPN achieved 0.63 IoU and 0.77 F1-score in test set. U-Net
achieved best accuracy at 0.70 IoU and 0.82 F1-score. LinkNet had worst
accuracy with only 0.45 IoU and 0.62 F1-score. We therefore chose U-Net
to inference as it has best accuracy in the test set. Figure \ref{fig:2.16} shows the
example of segmentation.

\begin{table}[H]
  \centering
  \caption{Test accuracy for architectures}
  \label{tab:2.2}
  \begin{tabular}{|p{3cm}||p{3cm}|p{3cm}|}
    \hline
    \multicolumn{3}{|c|}{Test accuracy} \\ \hline
    Architecture & Mean IoU & Mean F1-score \\ \hline
    U-Net   & 0.70 & 0.82 \\
    FPN     & 0.63 & 0.77 \\
    LinkNet & 0.45 & 0.62 \\ \hline
  \end{tabular}
\end{table}

\section{Comparison of YOLOv5s and Custom Segmentation Model}

When measuring inference time of YOLOv5s and custom segmentation model,
we observed YOLOv5s model from Ultralytics performed worse than our custom
segmentation model, as presented in Table \ref{tab:2.3}. There are few reasons
why the model performed worse that semantic segmentation. Firstly is that
we have not used lightest version of YOLOv5 model, which is YOLOv5n.
Secondly, we used 248 images to train, test and validate object detection,
whereas we only used 33 images for semantic segmentation. This, again, can
cause slowdown for YOLOv5s model. Lastly, YOLOv5s uses 640x640 images
to detect classes, whereas semantic segmentation uses 320x320 images. Since
YOLOv5s uses higher resolution to detect images, it could be sped up if it
had layers to input lower resolution. In the next section, we investigated the
second cause.

\begin{table}[H]
  \centering
  \caption{Comparison of inference time between YOlOv5s and custom segmentation}
  \label{tab:2.3}
  \begin{tabular}{|p{1.5cm}||p{3cm}|}
    \hline
    \multicolumn{2}{|c|}{Comparison of inference time} \\ \hline
    Model & Inference time (s) \\ \hline
    YOLOv5s & 1.09 \\
    Custom segmentation & 0.43 \\ \hline
  \end{tabular}
\end{table}

\section{Dataset Impact on YOLOv5s}
\label{sec:2.5}

We changed image dataset to only have 33 images and removed head category.
The number of classes was also reduced to 1. The dataset contains
training, validation, testing sets with 20, 5, 8 images, similar to the custom
dataset in semantic segmentation. We retrained our model with 100 epochs
with batch size of 4.

\begin{figure}[H]
  \centering
  \includegraphics[width=13cm]{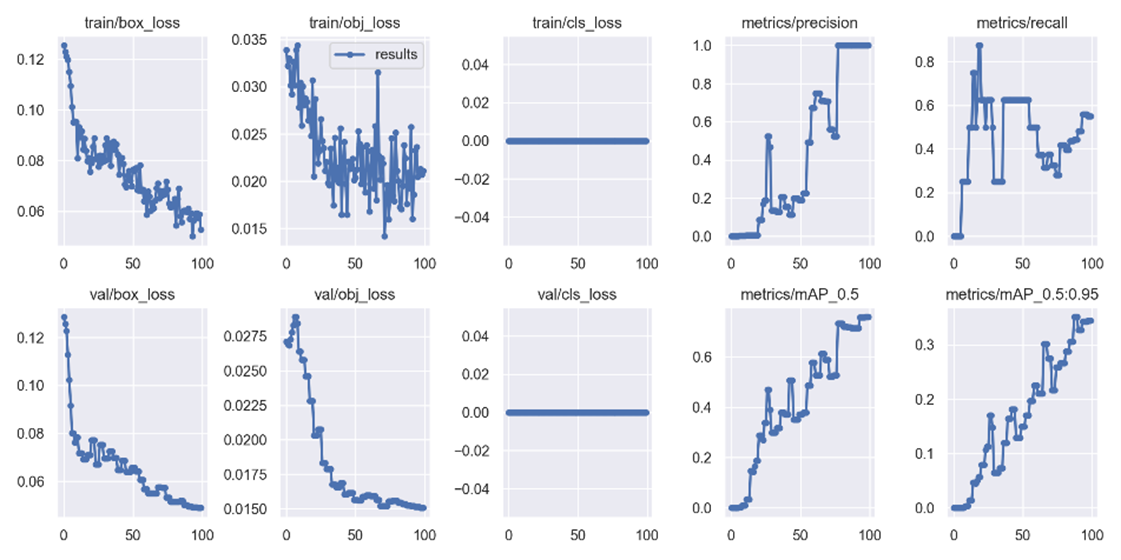}
  \caption{Loss and accuracy}
  \label{fig:2.17}
\end{figure}

\begin{figure}[H]
  \centering
  \includegraphics[width=10cm]{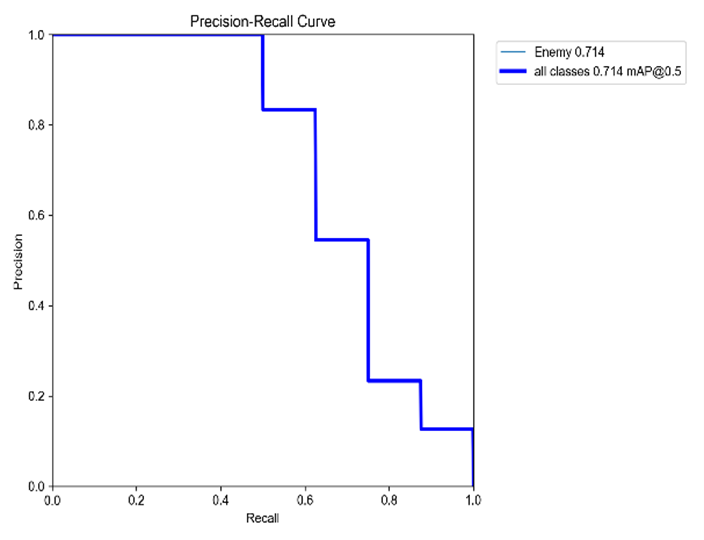}
  \caption{Precision to recall}
  \label{fig:2.18}
\end{figure}

\begin{figure}[H]
  \centering
  \includegraphics[width=10cm]{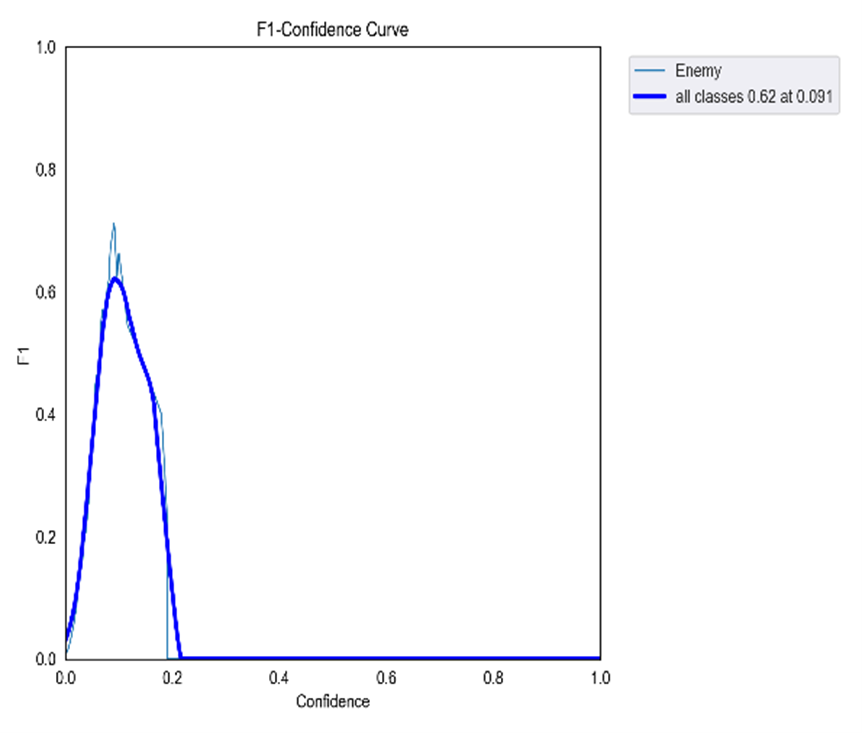}
  \caption{F1 score}
  \label{fig:2.19}
\end{figure}

\begin{figure}[H]
  \centering
  \includegraphics[width=10cm]{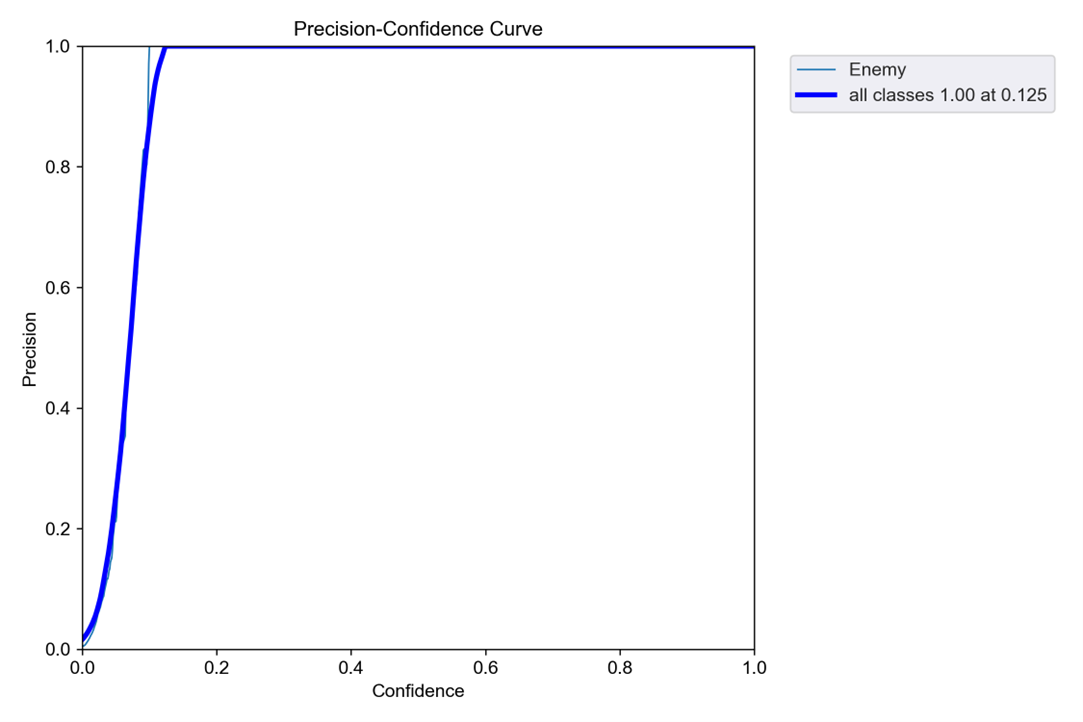}
  \caption{Precision to confidence}
  \label{fig:2.20}
\end{figure}

\begin{figure}[H]
  \centering
  \includegraphics[width=10cm]{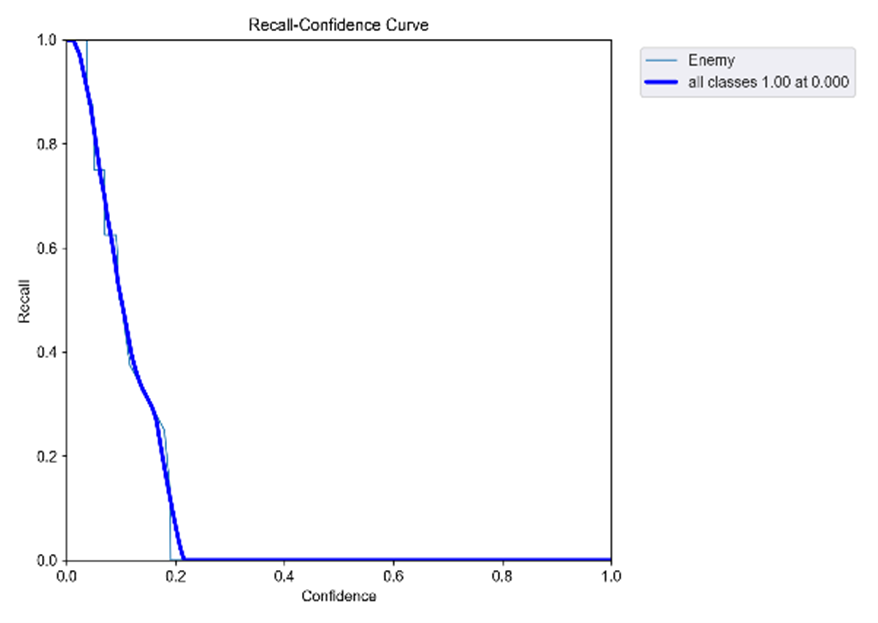}
  \caption{Recall to confidence}
  \label{fig:2.21}
\end{figure}

\begin{figure}[H]
  \centering
  \includegraphics[width=4cm]{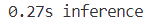}
  \caption{Inference time with a small dataset}
  \label{fig:2.22}
\end{figure}

Figure \ref{fig:2.17} shows the loss throughout the training. Because we used a
smaller dataset, we observe the loss was fluctuating more to learn the correct
parameters. We also see that cls loss is constant. This is because we were
using only one class. The area under precision to recall curve significantly
increased, as illustrated in Figure \ref{fig:2.18}. After the change, we see that the
peak of F1-score has shifted more to the left at 0.091, as depicted in Figure
\ref{fig:2.19}. We noticed that using a smaller dataset decreased the confidence to
achieve higher accuracy. Nevertheless, it is still able to predict very accurate
results with 0.62 F1-score at peak. This is because we are only predicting
one category. The same shift to the left occured in precision to confidence
and recall to confidence curve in Figures \ref{fig:2.20} and \ref{fig:2.21}. Now precision gets
high faster if the confidence slightly increases. In contrast, recall plunges
if the confidence increases. Ultimately, we observe that the inference time
has dropped to approximately 0.27 seconds in Figure \ref{fig:2.22}. This is a huge
improvement compared to our previous model and slightly faster than semantic
segmentation. Notice that we trained our model with the batch size
of 4, which is lower than the batch size used in semantic segmentation. \cite{jocher2020yolov5}
has made a study to compare the batch size and its corresponding inference
time, and concluded that higher batch size could result in higher FPS, therefore
lower inference time. Remember, we have not yet used YOLOv5n and
modified the layers of the model to take 320x320 inputs. Hence, YOLOv5s
was at disadvantage but still managed to outperform semantic segmentation
in terms of speed.

\subsection{Class Impact on Inference Time}

\begin{figure}[H]
  \centering
  \includegraphics[width=4cm]{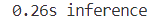}
  \caption{Inference time with a small dataset but 2 classes}
  \label{fig:2.23}
\end{figure}

In order to investigate the impact of dataset size and class number on the
inference time of YOLOv5s, we obtained results from the experiment in \ref{sec:2.5}.
Our initial findings indicate that using a smaller dataset and fewer classes
can significantly reduce inference time. However, it remains unclear whether
this effect is due to the smaller dataset size or the reduced number of classes.
To further explore this question, we trained a YOLOv5s model using a small
dataset of 33 images but with two classes instead of one. Our focus in this
experiment was on the inference time, which is shown in Figure \ref{fig:2.23}. We see
on the figure that the inference did not have a significant change as inference
time was 0.26 seconds. We can now conclude that the size of the dataset can
change the inference time but the number of classes does not have impact
on inference time. The additional results of this experiment are presented in
the Appendix \ref{app:B}, along with detailed graphs.

\section{Maskformer (Panoptic Segmentation)}

\begin{figure}[H]
  \centering
  \includegraphics[width=13.3cm]{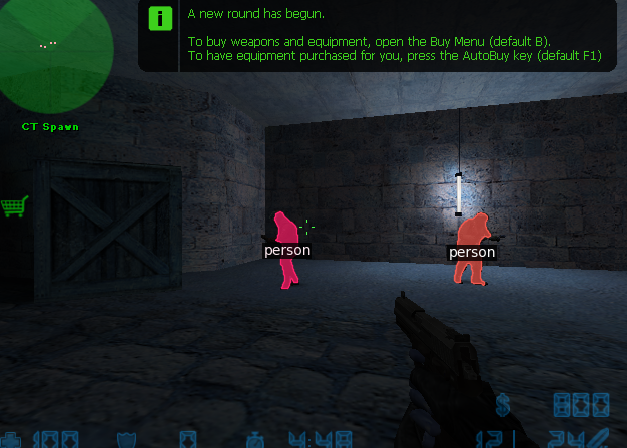}
  \caption{Maskformer panoptic segmentation}
  \label{fig:2.24}
\end{figure}

\begin{figure}[H]
  \centering
  \includegraphics[width=4cm]{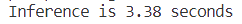}
  \caption{Inference time of Maskformer}
  \label{fig:2.25}
\end{figure}

We now take a look at Maskformer which is model that is able to predict
panoptic segmentation \cite{kirillov2019panoptic} of an image. The difference between panoptic
segmentation and semantic segmentation is that after when we segment objects,
we also try to differentiate instances. In addition, it gives a new metric
called panoptic quality (PQ) to measuring performance in a unified manner.
One such model is Maskformer \cite{cheng2021maskformer} developed by Facebook. It uses mask classification
to segment objects instead of per-pixel classification. This means
that instead of assigning each pixel to classes, it tries to create a binary
mask for each class from an image. This simplifies the segmentation task
and can infer both semantic and panoptic outputs. \cite{cheng2021maskformer} had observed the
accuracy of this model on COCO Panoptic Segmentation dataset achieved
PQ of 46.5. COCO dataset is large scale project by Microsoft where it already
has large number of labeled segmentation images and is often used for
benchmarking accuracy \cite{lin2014coco}. Similar projects like COCO include ADE20K
\cite{zhou2017ade20k}, \cite{zhou2018ade20k}. We will use pretrained version of this model on ADE20K dataset
to demonstrate our novel prototype. The weights are uploaded to Model
Zoo, and we are going to use the uploaded model with model id 42747488\_1
and R50 backbone, which is trained on ADE20K dataset. Images below are
segmented using panoptic segmentation and semantic segmentation. Figure
\ref{fig:2.24} shows the example of panoptic segmentation and Figure \ref{fig:2.25} illustrates
the inference time.

\chapter{Novel Architecture}

\begin{figure}[H]
  \centering
  \includegraphics[width=11cm]{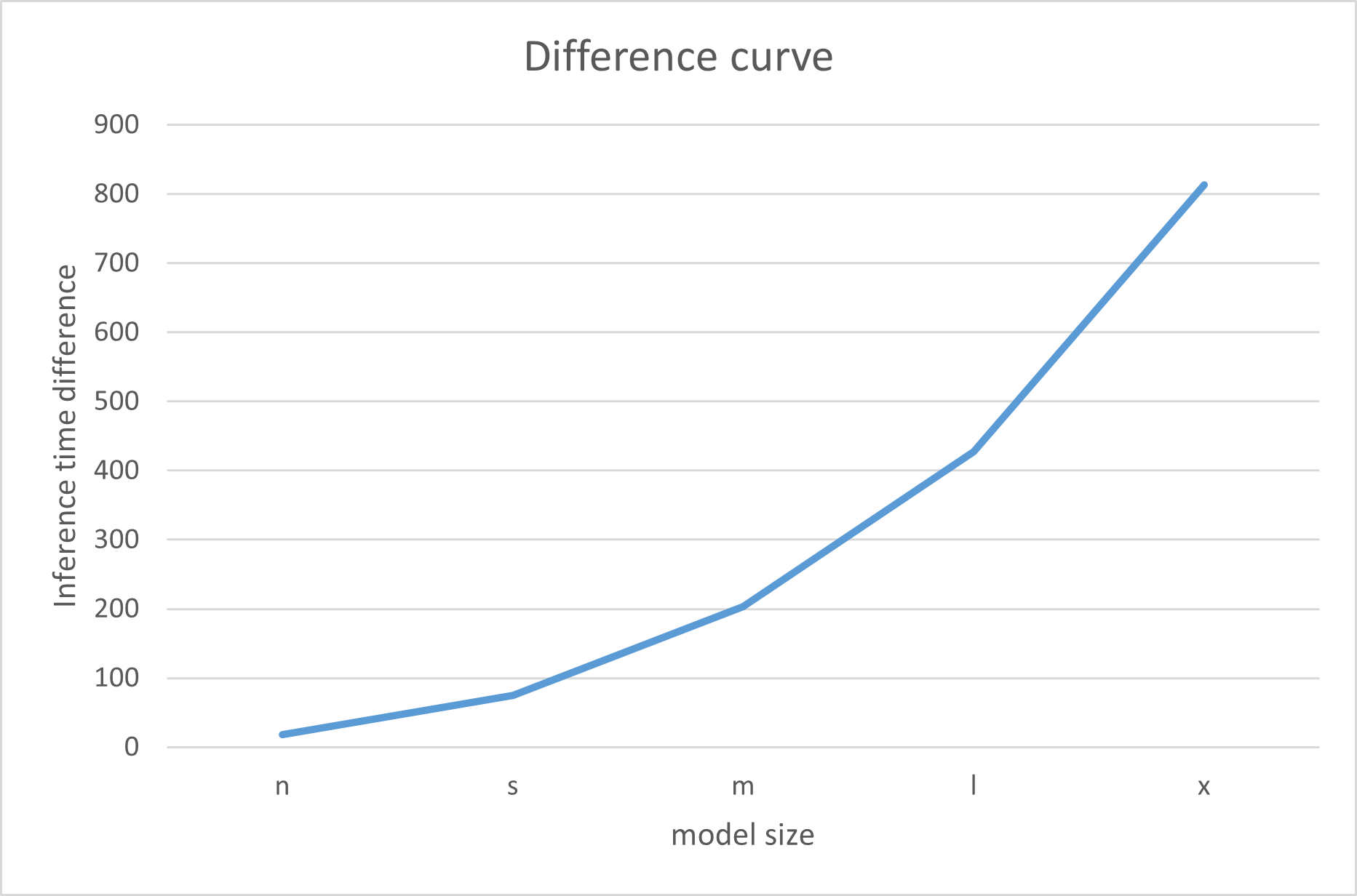}
  \caption{Exponential blowup}
  \label{fig:3.1}
\end{figure}

We have measured the inference time of YOLOv5s and the custom segmentation
model on our custom dataset, and found that YOLOv5s is faster than
segmentation, with 0.27 seconds and custom segmentation at 0.43 seconds.
We see that the difference of inference time is not significant. However, one
interesting question remains: does the discrepancy in inference time between
object detection and semantic segmentation grow when using more complex
model? Ultralytics \cite{jocher2020yolov5} conducted experiments with both segmentation and
object detection algorithms. Their lightest object detection model was measured
on a CPU with 0.0450 seconds, while their largest model had 0.7660
seconds. For segmentation, their lightest model had 0.0627 seconds, while
their largest model had 1.5792 seconds. We can see that the light segmentation
model was 1.4 times slower than the light object detection model, but
the large segmentation model was 2.0 times slower. This shows that the
difference grows over time. We have plotted the graph of the difference from
the experiment, which is shown in Figure \ref{fig:3.1}. We observe from the figure
that the growth is exponential, which is a huge cost for a model. The letters
on the x-axis are the complexities of models and on the y-axis is the inference
time of models. As models get more complex, the (exponentially) longer it
takes to infer images. Thus, we need to find a solution with the best of two
worlds.

\section{Adaptive Distance Detection}
\label{sec:3.1}

\subsection{Design}

\begin{figure}[H]
  \centering
  \includegraphics[width=6cm]{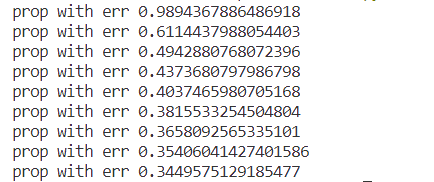}
  \caption{Decreasing proportional difference}
  \label{fig:3.2}
\end{figure}

The adaptive distance detection takes the advantage of the speed of object
detection and the precision of semantic segmentation. Generally, if we do
not care about speed, we would use semantic segmentation all the time.
However, this is not the case here as reaction speed is very important in FPS
shooting games. The idea of adaptive distance detection is following. If the
player is near, use object detection. If the player is far away, use semantic
segmentation. The reason why we choose object detection when player is
based on the following argument. Assume there is a circle, which will be
our target object to detect using object detection algorithm. The detected
bounding box will be a square with length same as diameter of circle. Now,
assume there exists a small error on the algorithm so that our bounding box
does not fit exactly to the circle. We then calculate the difference between
bounding box with error and circle. Finally, we argue that the proportion
of the difference with regards to circle gets smaller as our target circle gets
larger. This property can be shown by Listing \ref{lst:3.1} and Figure \ref{fig:3.2}.

\begin{lstlisting}[style=py, caption={Code snippet demonstrating decreasing proportional difference while increasing circle size}, label={lst:3.1}]
import math
error = 0.5
for i in range(1,10):
    radius = i
    circle = math.pi  * i**2
    diame = 2 * radius + error
    rect = diame*diame
    diff = abs(rect - circle)
    print("prop with difference",diff / circle)
\end{lstlisting}

\begin{figure}[H]
  \centering
  \includegraphics[width=9cm]{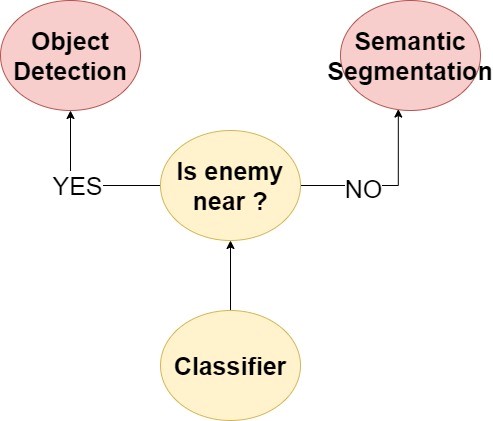}
  \caption{Adaptive distance detection}
  \label{fig:3.3}
\end{figure}

We note that this property is helpful in an FPS game. We might get a
detection error from object detection but still able to end up aiming at enemy
body when enemy is near because of decreasing proportional difference. And
because of this, it becomes obvious to use object detection when enemy is
near to not only get performance speed but also tolerance to error. However,
we still need to think about the bridge between object detection and semantic
segmentation. Our solution to this problem is to add an additional classifier
to the algorithm. The classifier model will attempt to predict instances where
enemy is nearby. If so, we switch on object detection algorithm. Otherwise,
we use semantic segmentation (Figure \ref{fig:3.3}). With this bridge we can benefit
both performance speed from object detection algorithm when enemy is near
and accuracy from semantic segmentation when enemy is far.

\subsection{Dataset for Classifier}

Our dataset for classifier contains 102 images. We split it to train, validation,
test sets just like in semantic segmentation and object detection. However,
it was not enough to prevent overfitting. To solve this, we also implemented
generated data with augmentations so that the model can generalize more.
We implemented horizontal flip, rotation and brightness change for augmentations.

\subsection{Custom Classifier}

We trained custom classifier to predict instances where enemy is nearby.
Our own dataset was created and contained 2 classes (far, near). Model was
customized and had following layers presented in Listing \ref{lst:3.2}. A Conv2D is
added to our model, in which it has 32 3 x 3 kernels with ReLU \cite{agarap2019relu} activation
function. Then a we added a MaxPooling layer of size 2. After that, we
flatten our data to 1 dimensional vector and pass it to densel layer. The
dense layer is an interchangeable term with hidden layer. We use ReLU \cite{agarap2019relu}
activation in the first dense layer. Sigmoid \cite{pratiwi2020sigmoid} activation function is used in
second layer. Finally, we compile our model with Adam optimizer \cite{kingma2017adam} and
binary crossentropy loss function \cite{ruby2020bce}. We train our model with l1 regularizer
\cite{tibshirani1996lasso}, batch size of 8. The custom model is compared to AlexNet \cite{krizhevsky2012alexnet}, which
only consists of 8 layers, using accuracy metrics. We used accuracy as our
metric instead of F1-score and IoU because the negatives are as important
as positives to detect whether enemy is far, therefore having similar cost.

\begin{lstlisting}[style=py, caption={Model layers for adaptive distance detection classifier}, label={lst:3.2}]
import tensorflow as tf
model = tf.keras.Sequential()
model.add(tf.keras.layers.Conv2D(32, kernel_size=3,
                                 activation='relu',
                                 input_shape=(480, 642, 3)))
model.add(tf.keras.layers.MaxPooling2D(pool_size=2))
model.add(tf.keras.layers.Flatten())
model.add(tf.keras.layers.Dense(64, activation='relu'))
model.add(tf.keras.layers.Dense(1, activation='sigmoid'))
model.compile(optimizer='adam', loss='binary_crossentropy',
              metrics=['accuracy'])
\end{lstlisting}

\subsection{Results of Classifier}

\begin{figure}[H]
  \centering
  \includegraphics[width=12cm]{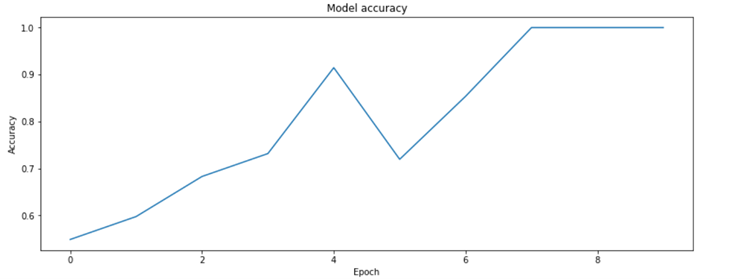}
  \caption{Accuracy of custom classifier}
  \label{fig:3.4}
\end{figure}

\begin{figure}[H]
  \centering
  \includegraphics[width=12cm]{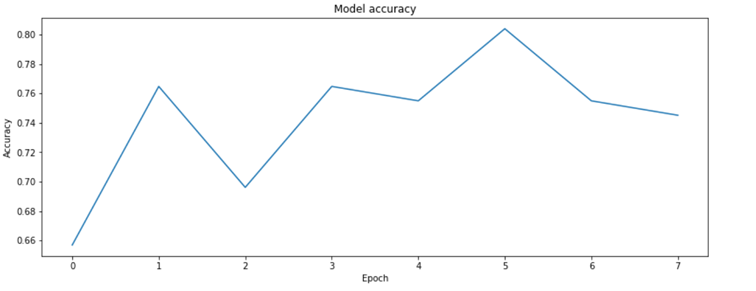}
  \caption{First augmentations of custom classifier}
  \label{fig:3.5}
\end{figure}

\begin{figure}[H]
  \centering
  \includegraphics[width=11.8cm]{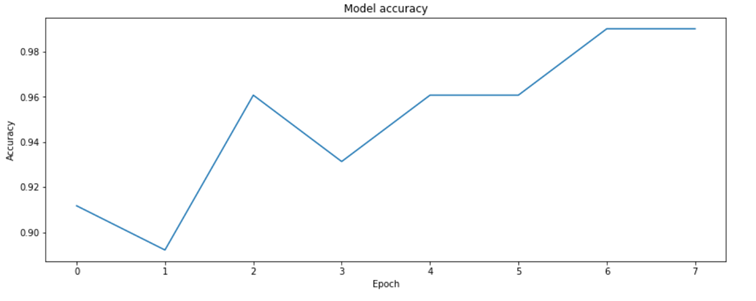}
  \caption{Second augmentations of custom classifier}
  \label{fig:3.6}
\end{figure}

\begin{figure}[H]
  \centering
  \includegraphics[width=12cm]{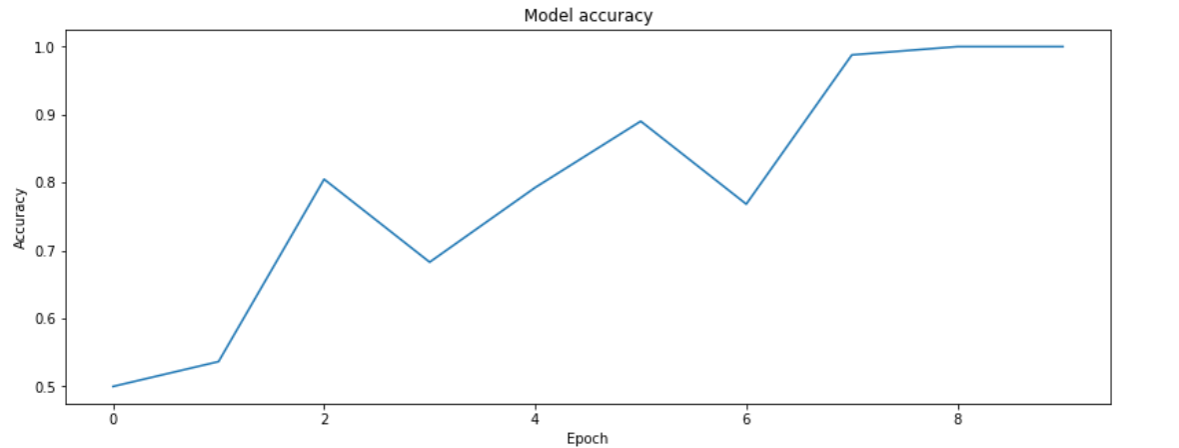}
  \caption{Accuracy of AlexNet}
  \label{fig:3.7}
\end{figure}

\begin{figure}[H]
  \centering
  \includegraphics[width=12cm]{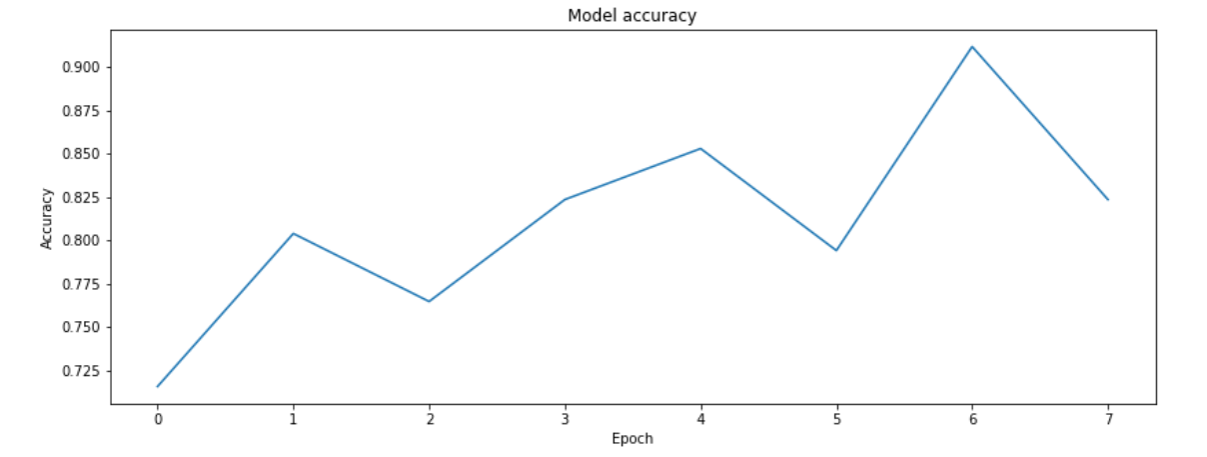}
  \caption{First augmentations of AlexNet}
  \label{fig:3.8}
\end{figure}

\begin{figure}[H]
  \centering
  \includegraphics[width=13cm]{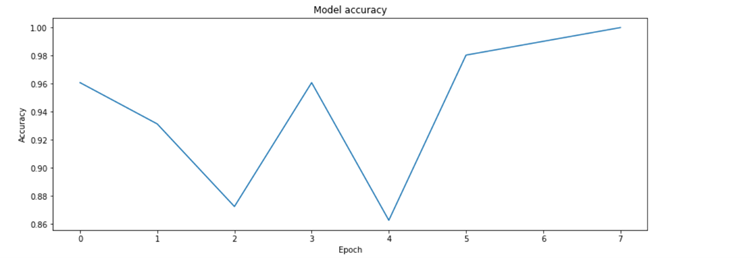}
  \caption{Second augmentations of AlexNet}
  \label{fig:3.9}
\end{figure}

Figure \ref{fig:3.4} shows the accuracy of our custom classifier throughout training.
Compared to Figure \ref{fig:3.7}, we see both models reached 100 percent accuracy
with just within 10 epochs. During first augmentation, both models performed
worse, as depicted in Figures \ref{fig:3.5} and \ref{fig:3.8}. Nevertheless, Alexnet
\cite{krizhevsky2012alexnet} had slightly better accuracy. During second augmentation, we see that
AlexNet was very unstable (Figure \ref{fig:3.9}). It had worse accuracy than our custom
classifier in Figure \ref{fig:3.6}. Finally, during testing both models performed
very well, as shown in Table \ref{tab:3.1}. However, considering that AlexNet \cite{krizhevsky2012alexnet}
was not stable during second augmentation, we decided to use our custom
model as a bridge as it was more stable.

\begin{table}[H]
  \centering
  \caption{Test accuracy for classifiers}
  \label{tab:3.1}
  \begin{tabular}{|p{3cm}||p{3cm}|}
    \hline
    \multicolumn{2}{|c|}{Test accuracy} \\ \hline
    Classifier & Mean accuracy \\ \hline
    Custom  & 1.00 \\
    AlexNet & 1.00 \\ \hline
  \end{tabular}
\end{table}

\subsection{Implementation of Adaptive Distance Detection}
\label{sec:3.1.5}

\begin{figure}[H]
  \centering
  \includegraphics[width=12cm]{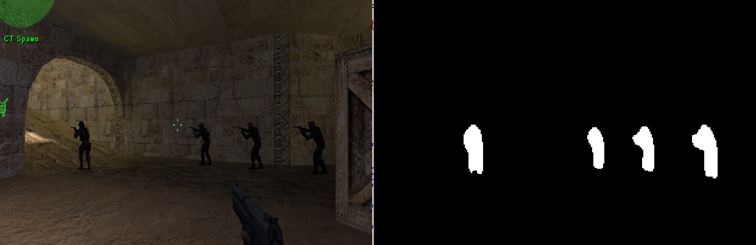}
  \caption{Segmentation when enemy is far}
  \label{fig:3.10}
\end{figure}

\begin{figure}[H]
  \centering
  \includegraphics[width=12cm]{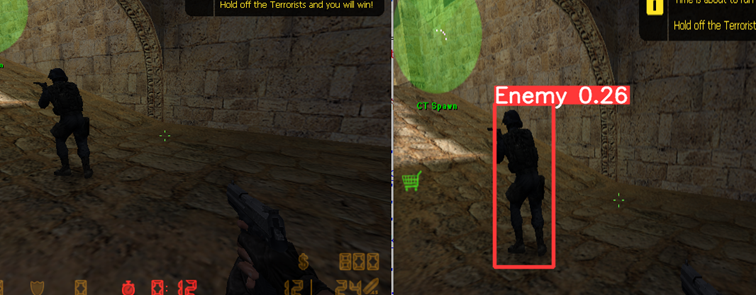}
  \caption{Object detection when enemy is near}
  \label{fig:3.11}
\end{figure}

We now have all the essential materials and just need implementation to
connect semantic segmentation, object detection and classification. Trained
models from Chapter 2 (YOLOv5s, custom segmentation model with U-net
architecture) were combined with the classifier. All model weights were
downloaded from drive and added to our project space. Since we do not
have an NVIDIA graphics card, we ensured that all models were configured
to use the CPU for inference. Our code space resembles as a project
monorepo with imported and cloned external repositories, such as Maskformer
\cite{maskformerrepo} and YOLOv5s \cite{jocher2020yolov5}. We implemented further code on top of this
foundation, extensively utilizing libraries such as Pytorch \cite{paszke2019pytorch} and Tensorflow
\cite{abadi2015tensorflow}. Additional libraries were used in our project and can be found in
the requirements file. We deeply acknowledge and thank for the contributions
from them. After implementing our architecture, we obtained Figures
\ref{fig:3.10} and \ref{fig:3.11}.

\section{Adaptive Instance Segmentation}

\subsection{Design}

\begin{figure}[H]
  \centering
  \includegraphics[width=9cm]{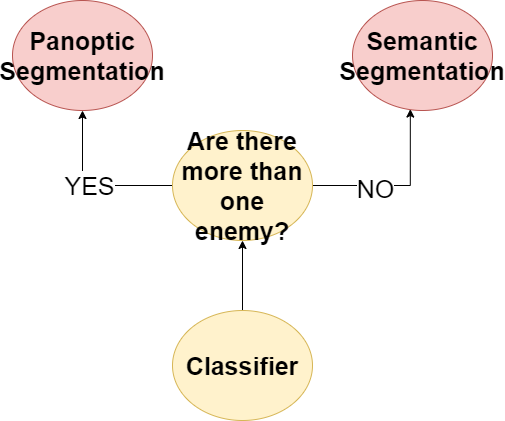}
  \caption{Design of adaptive instance segmentation}
  \label{fig:3.12}
\end{figure}

Much like adaptive distance detection, it is possible to apply the same concept
to semantic segmentation and panoptic segmentation. Panoptic segmentation
differentiates instances from images, but it could be an overkill
to use panoptic segmentation when there is only one instance on the screen.
This encourages the proposal of using semantic segmentation when there
is one instance and panoptic segmentation when there are more instances.
Figure \ref{fig:3.12} illustrates the design of such architecture.

\subsection{Custom Classifier and its Dataset}

Classifier was used as a bridge between them and was trained on a custom
dataset with 350 images that was split to train, validation, test sets.
Our classifier has the same layers as the custom model used in Section \ref{sec:3.1}.
Augmentation images were also used to generalize the model to prevent overfitting.

\subsection{Results of Classifier}

\begin{figure}[H]
  \centering
  \includegraphics[width=12cm]{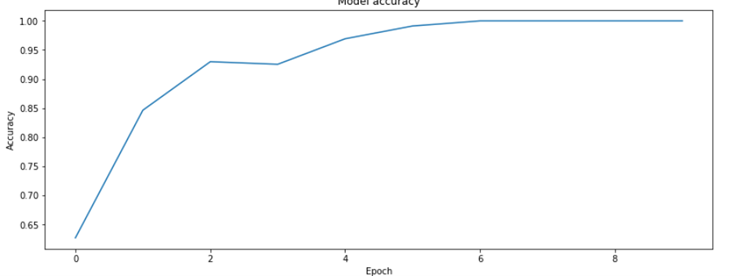}
  \caption{Accuracy of custom classifier}
  \label{fig:3.13}
\end{figure}

\begin{figure}[H]
  \centering
  \includegraphics[width=12cm]{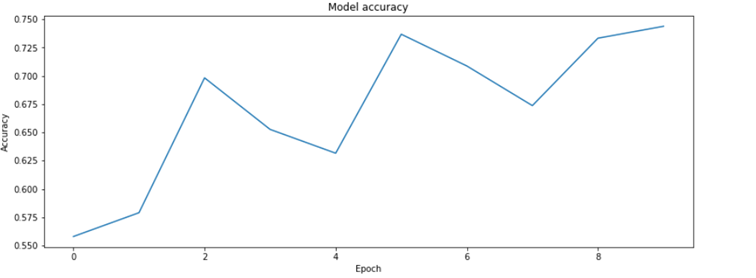}
  \caption{First augmentations of custom classifier}
  \label{fig:3.14}
\end{figure}

\begin{figure}[H]
  \centering
  \includegraphics[width=12cm]{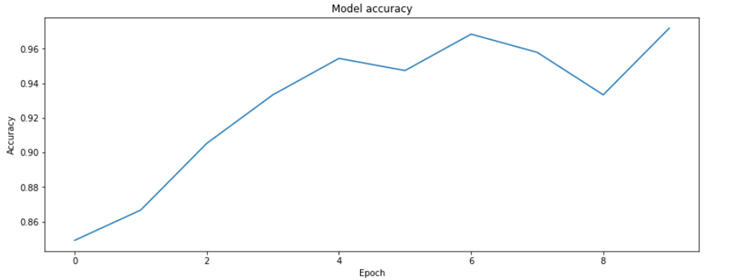}
  \caption{Accuracy of second augmentations}
  \label{fig:3.15}
\end{figure}

\begin{figure}[H]
  \centering
  \includegraphics[width=12cm]{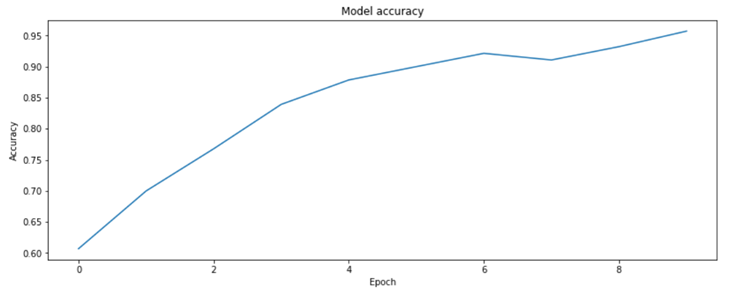}
  \caption{Accuracy of AlexNet}
  \label{fig:3.16}
\end{figure}

\begin{figure}[H]
  \centering
  \includegraphics[width=13cm]{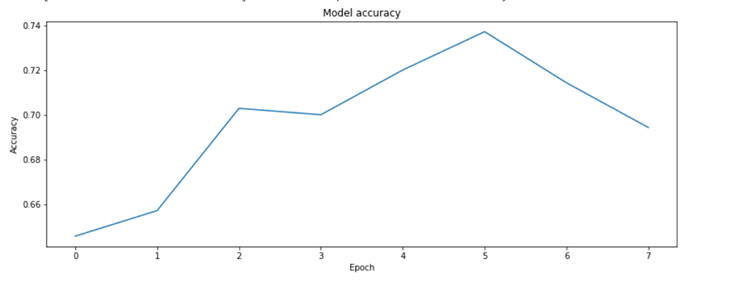}
  \caption{First augmentations of AlexNet}
  \label{fig:3.17}
\end{figure}

\begin{figure}[H]
  \centering
  \includegraphics[width=12cm]{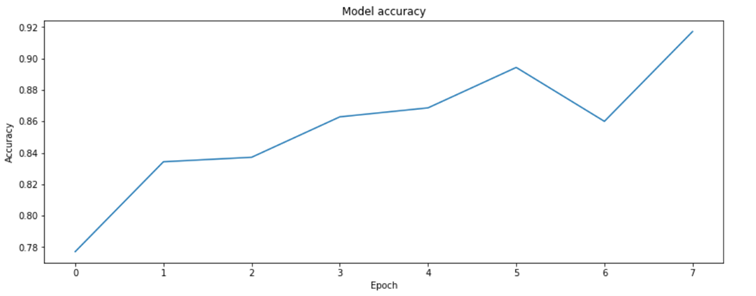}
  \caption{Accuracy of second augmentations}
  \label{fig:3.18}
\end{figure}

The accuracy of both custom and AlexNet are shown in Figures \ref{fig:3.13} and
\ref{fig:3.16}. We observe that our custom model reached to the global minimum
faster than AlexNet. During first augmentations, our custom model was
unstable, but reaching good performance at the end, as depicted in Figure
\ref{fig:3.14}. In contrast, AlexNet was stable at first, but then getting unstable
when reaching the end of epochs, as presented in Figure \ref{fig:3.17}. Figures \ref{fig:3.15}
and \ref{fig:3.18} show the accuracy of both models during second augmentations.
Again, we observe our model was getting high accuracy faster than AlexNet.
Finally, both models performed well during testing but our custom classifier
did slightly better. This is represented in Table \ref{tab:3.2}. Therefore, we chose to
implement adaptive instance segmentation with our custom classifier.

\begin{table}[H]
  \centering
  \caption{Test accuracy for classifiers}
  \label{tab:3.2}
  \begin{tabular}{|p{3cm}||p{3cm}|}
    \hline
    \multicolumn{2}{|c|}{Test accuracy} \\ \hline
    Classifier & Mean accuracy \\ \hline
    Custom  & 0.97 \\
    AlexNet & 0.78 \\ \hline
  \end{tabular}
\end{table}

\subsection{Implementation of Adaptive Instance Segmentation}

\begin{figure}[H]
  \centering
  \includegraphics[width=10cm]{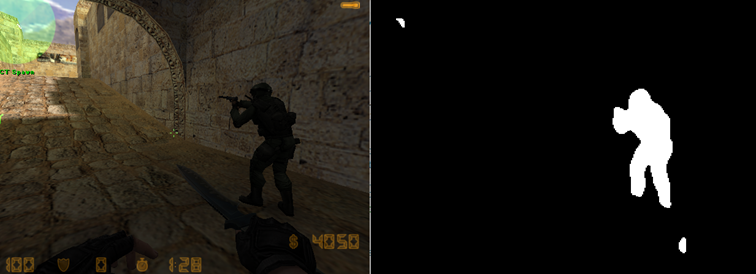}
  \caption{Semantic segmentation if there is one instance}
  \label{fig:3.19}
\end{figure}

\begin{figure}[H]
  \centering
  \includegraphics[width=10cm]{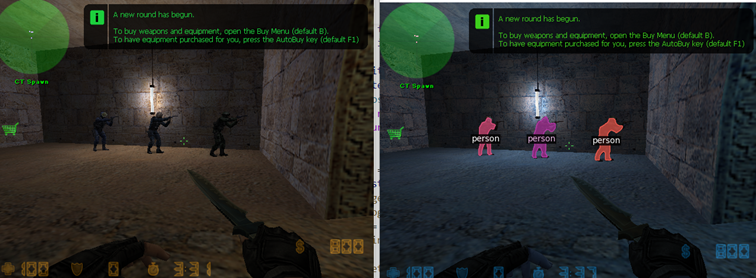}
  \caption{Panoptic segmentation if there are more than one instance}
  \label{fig:3.20}
\end{figure}

After training classifier, we implemented adaptive instance segmentation to
connect our custom classifier, custom segmenation model and panoptic model
from Maskformer. Most of the tools used are already mentioned in \ref{sec:3.1.5}.
Maskformer \cite{maskformerrepo} takes BGR images as input, so it is needed to convert RGBA
format when taking screenshots to BGR format if we are using panoptic
segmentation. We obtained Figures \ref{fig:3.19} and \ref{fig:3.20}.

\chapter{Evaluation}

\section{Testing of Adaptive Distance Detection}
\label{sec:4.1}

It was a challenge to think of an appropriate testing metric for adaptive distance
detection. Hypothetically, we believe that there is approximately 50 \%
chance that the enemy could be close and 50 \% that enemy could be far. Initially,
we thought of a `randomized testing approach'. This approach takes
an assumption that there 50\% chance that the enemy is close and chooses
object detection and semantic segmentation in a randomized manner. However,
this approach is flawed in a sense that it could take a picture in which
the enemy is far away and use object detection. We then used `fair testing'
approach, which we explicitly define when to use object detection (if enemy
is near) or semantic segmentation (if enemy is far away). In fact, this will
simulate a perfect classifier that will output 1 when enemy is near, 0 when
enemy is far. It is infinitely fast. A specific dataset consisting of 10 images
was created for testing. For the ground truth labels, segmentation annotations
were used. A thorough comparison of adaptive distance detection,
semantic segmentation, object detection was conducted.

Table \ref{tab:4.1} shows the comparison of adaptive distance detection to other
models. If we used object detection only, the mean IoU and F1-score are
0.18 and 0.25 respectively. This is expected since we are using segmentation
ground truth labels which makes object detection really bad at detecting
exact shapes. Its inference time is the fastest with 0.25 seconds. Semantic
segmentation is the most accurate out of three with mean IoU and F1-score
0.56 and 0.70 but slowest in terms of inference time with 0.48 seconds. If we
use adaptive distance detection, the accuracy and inference time are balanced
with each other, resulting in 0.36 IoU, 0.52 F1-score, 0.42 inference time.

\begin{table}[H]
  \centering
  \caption{Comparison of adaptive distance detection and other models}
  \label{tab:4.1}
  \begin{tabular}{|p{3cm}||p{3cm}|p{3cm}|p{3cm}|}
    \hline
    \multicolumn{4}{|c|}{Test accuracy and inference time} \\ \hline
    Model & Mean IoU & Mean F1-score & Mean inference (s) \\ \hline
    Adaptive distance detection & 0.36 & 0.52 & 0.42 \\
    Object detection (YOLOv5s) & 0.18 & 0.25 & 0.28 \\
    Custom semantic segmentation & 0.56 & 0.70 & 0.48 \\ \hline
  \end{tabular}
\end{table}

\section{Testing of Adaptive Instance Segmentation}

The same fair testing technique introduced in Section \ref{sec:4.1} is used in adaptive
instance segmentation. The dataset contains 8 images, with half having
only instance of enemy, and the other half having more instances of enemies.
One thing to note is that Maskformer (panoptic segmentation) was
pretrained on ADE20K, which is a large dataset. This ultimately makes the
model more complex than our custom segmentation model, therefore having
longer inference time, but better accuracy. This encourages the benchmarking
of two equally complex models of semantic segmentation and panoptic
segmentation on further research.

\begin{table}[H]
  \centering
  \caption{Comparison of adaptive instance segmentation to other models}
  \label{tab:4.2}
  \begin{tabular}{|p{3cm}||p{3cm}|p{3cm}|p{3cm}|}
    \hline
    \multicolumn{4}{|c|}{Test accuracy and inference time} \\ \hline
    Model & Mean IoU & Mean F1-score & Mean inference (s) \\ \hline
    Adaptive instance segmentation & 0.72 & 0.83 & 1.87 \\
    Panoptic segmentation & 0.75 & 0.85 & 3.83 \\
    Custom semantic segmentation & 0.68 & 0.81 & 0.51 \\ \hline
  \end{tabular}
\end{table}

Table \ref{tab:4.2} shows the comparison of adaptive instance segmentation to
other models. Panoptic segmentation has the best accuracy with 0.75 mean
IoU and 0.85 mean F1-score. It took 3.83 seconds to infer an image. Semantic
segmentation has the lowest accuracy with 0.68 (mean IoU) and 0.81 (mean
F1-score). It has the fastest inference with 0.51 seconds. Adaptive distance
segmentation balanced inference time with accuracy, having 0.72 mean IoU
and 0.83 mean F1-score. It took 1.87 seconds to infer an image.

\section{Real-time Classifier for Adaptive Methods}

During testing of adaptive distance detection and adaptive instance segmentation,
we assumed that there is a classifier that is perfect, which means it
will always give correct output and takes 0 inference time. However, in real
practice the classifier will have a small margin of error and will take some
time to infer image. In adaptive distance detection, using the same dataset
with real-time classifier attached took longer to infer than semantic segmentation.
Partially, this is because we used a much bigger dataset to train
classifier, making it more complex than object detection and semantic segmentation.
But we also know that, as model grows, semantic segmentation
will take exponentially longer than object detection. Therefore, it becomes
clear that when given a model with big complexity, we should use adaptive
method. This is also applicable to adaptive instance segmentation. We
attached real-time classifiers to our adaptive distance detection and adaptive
instance segmentation models. We noticed that both models had slower
inference speed as expected, presented in Table \ref{tab:4.3}. Mean IoU, F1-score
and inference time of adaptive distance detection are 0.41, 0.56 and 0.54
seconds respectively. Mean IoU, F1-score and inference time of adaptive
instance segmentation with attached classifier are 0.70, 0.82, 1.02 seconds
respectively.

\begin{table}[H]
  \centering
  \caption{Test accuracy and inference of adaptive algorithms with real-time classifier}
  \label{tab:4.3}
  \begin{tabular}{|p{3cm}||p{3cm}|p{3cm}|p{3cm}|}
    \hline
    \multicolumn{4}{|c|}{Test accuracy and inference time} \\ \hline
    Model & Mean IoU & Mean F1-score & Mean inference (s) \\ \hline
    Adaptive distance detection & 0.41 & 0.56 & 0.54 \\
    Adaptive instance segmentation & 0.70 & 0.82 & 1.02 \\ \hline
  \end{tabular}
\end{table}

\section{Real-time Observations of Adaptive Distance Detection}

In this section, we provide various highlighted scenarios to observe our
model's behaviour in real-time.

\subsection{Scenario 1: When player is holding knife}

\begin{figure}[H]
  \centering
  \includegraphics[width=12cm]{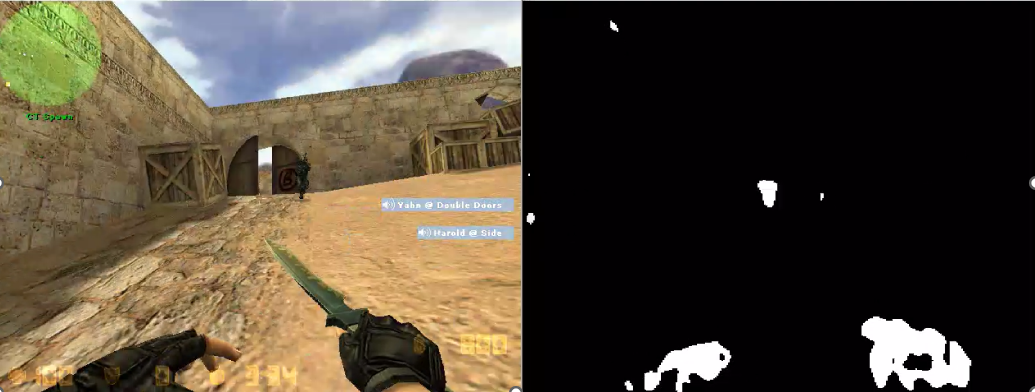}\\[-1pt]
  \includegraphics[width=12cm]{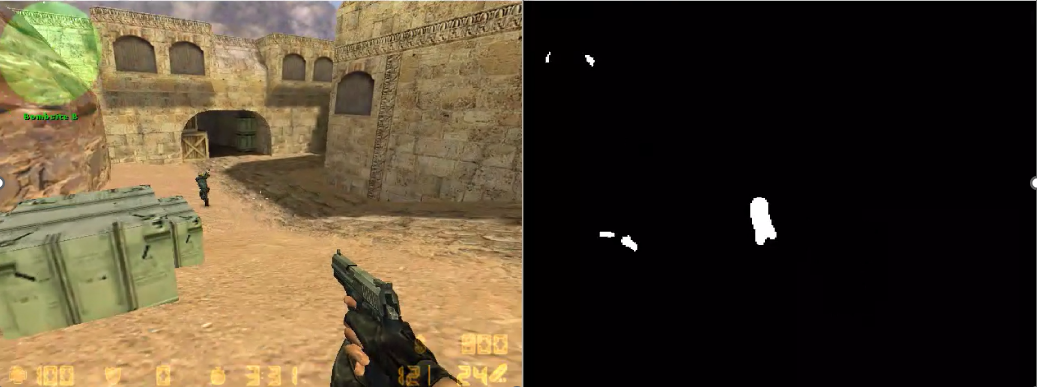}
  \caption{Semantic segmetation segments our knife. Top: With knife. Bottom: Without knife.}
  \label{fig:4.1}
\end{figure}

When adaptive distance detection uses semantic segmentation, the model
will confuse the player's hand as enemy when holding knife. This does not
happen when our player is holding a pistol. This is shown in Figure \ref{fig:4.1}.
The problem could be solved by improving our training dataset for custom
segmentation model by adding more diverse images, including player holding
knife.

\subsection{Scenario 2: Confusing Pistol as Enemy}

\begin{figure}[H]
  \centering
  \includegraphics[width=12cm]{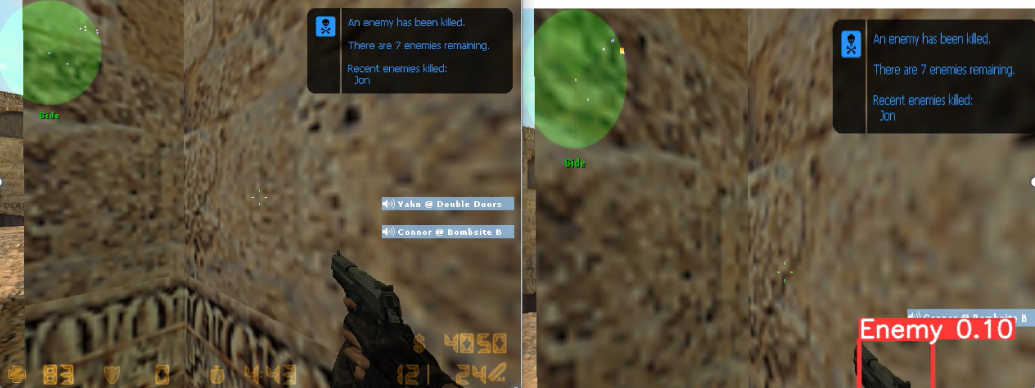}
  \caption{Object detection detecting our gun as enemy}
  \label{fig:4.2}
\end{figure}

Sometimes object detection confuses our pistol as enemy, as seen in Figure
\ref{fig:4.2}. This is caused by low connfidence when detecting images using
YOLOv5s. Again, this issue could be mitigated by adding larger data.

\subsection{Scenario 3: Multiple Enemies}

\begin{figure}[H]
  \centering
  \includegraphics[width=12cm]{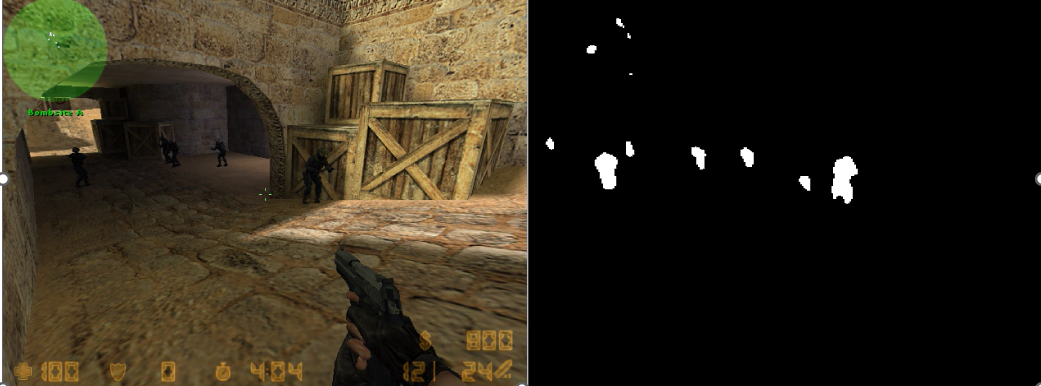}\\[-1pt]
  \includegraphics[width=12cm]{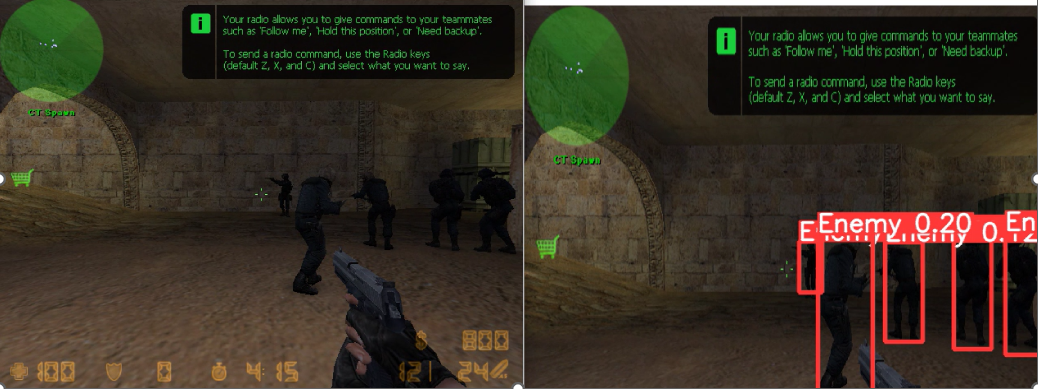}\\[-1pt]
  \includegraphics[width=12cm]{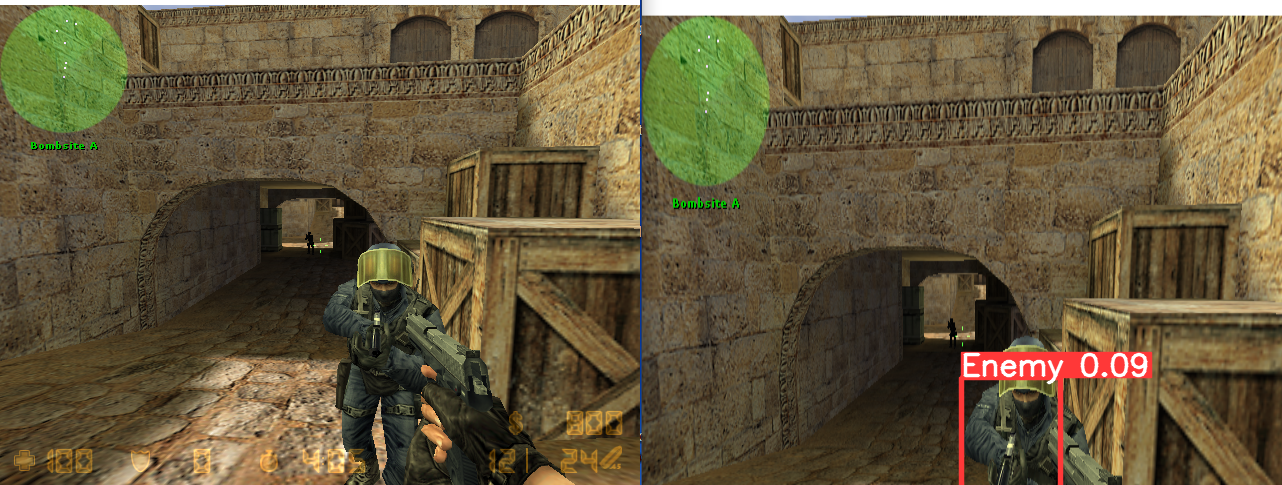}
  \caption{Adaptive distance detection detecting multiple enemies. Top: All enemies are far. Middle: All enemies are close. Bottom: Most enemies are far but one enemy is near.}
  \label{fig:4.3}
\end{figure}

If there are multiple enemies on the screen, and all of them are a far, then
our model will use semantic segmentation. If all enemies are near, our model
will use object detection. If there is at least one enemy on the screen that
is near, our model will use object detection. This is visually represented in
Figure \ref{fig:4.3}.

\subsection{Scenario 4: No Enemies}

\begin{figure}[H]
  \centering
  \includegraphics[width=12cm]{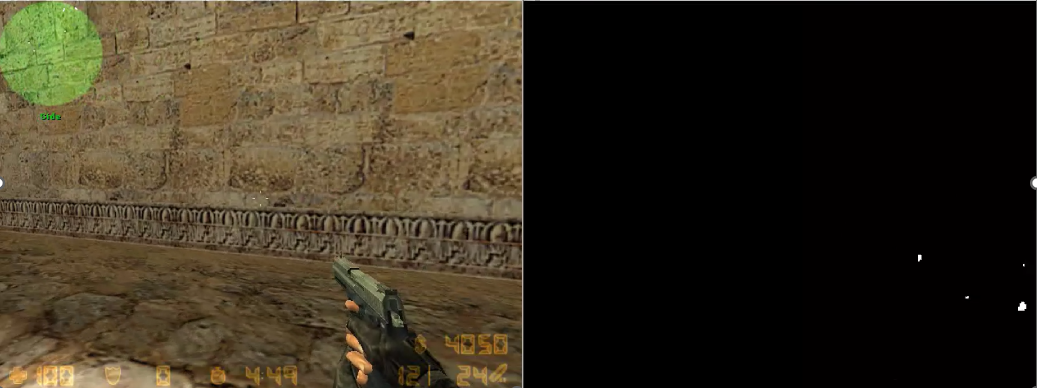}\\[-1pt]
  \includegraphics[width=12cm]{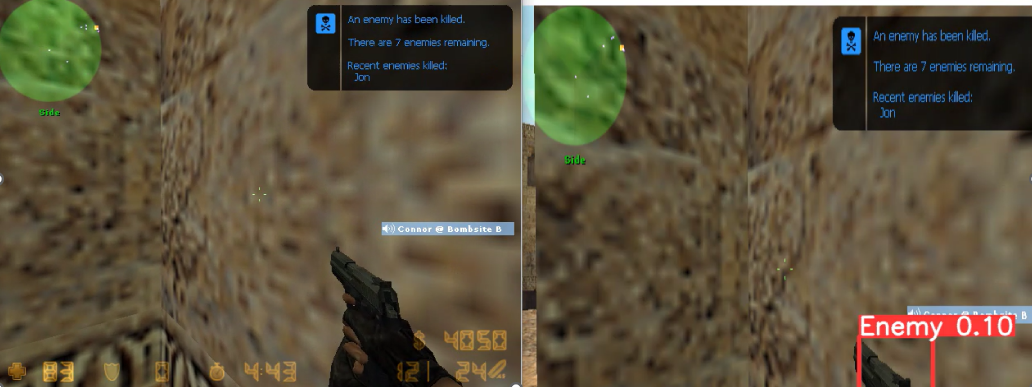}
  \caption{Randomized behaviour of adaptive distance detection}
  \label{fig:4.4}
\end{figure}

When there are no enemies, for example we are aiming at wall, the model
randomly chooses to use object detection or semantic segmentation, as can
be seen in Figure \ref{fig:4.4}. If we do not want to have this random behaviour, we
could add a new class that consists of images, which do not contain enemies
at all. This would mean to add a new set of data.

\subsection{Scenario 5: Finding the Boundary}

\begin{figure}[H]
  \centering
  \includegraphics[width=12cm]{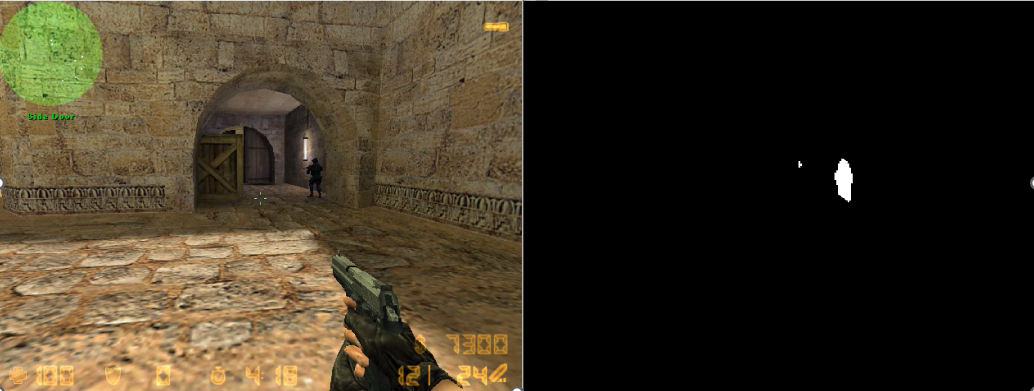}\\[-1pt]
  \includegraphics[width=12cm]{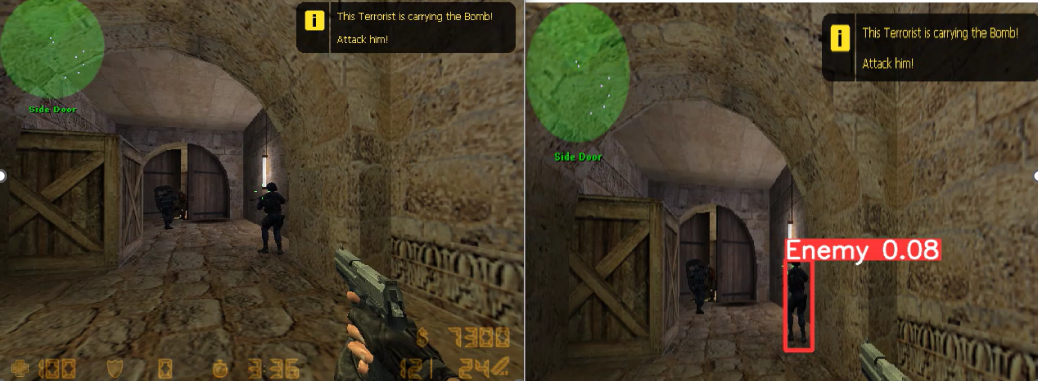}\\[-1pt]
  \includegraphics[width=12cm]{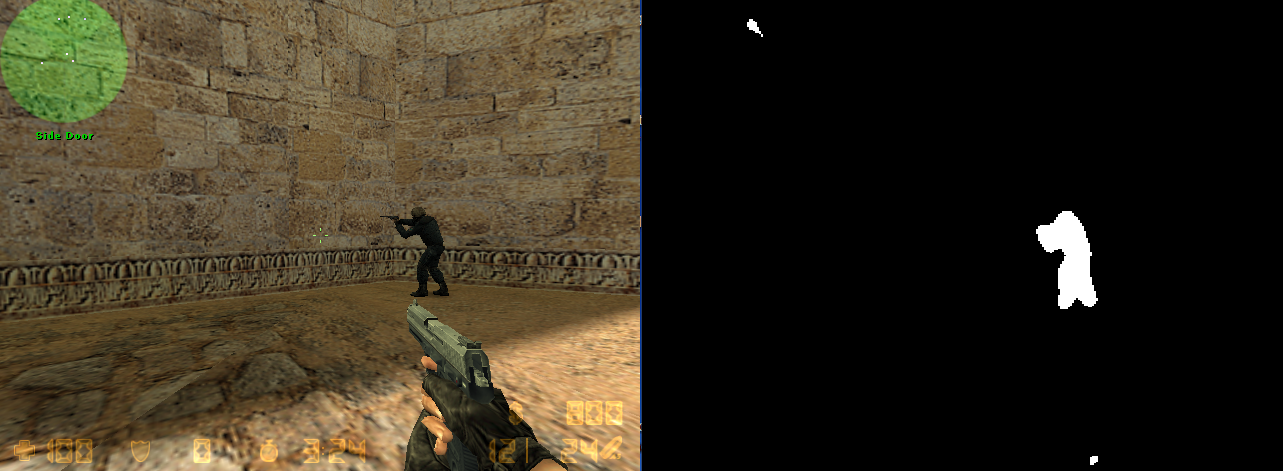}
  \caption{Boundary of adaptive distance detection. Top: Semantic segmentation distance. Middle: Object detection distance. Bottom: Error of boundary detection}
  \label{fig:4.5}
\end{figure}

The following figures show the boundary to switch from semantic segmentation
to object detection. However, it is possible that the classifier can
produce an error, using semantic segmentation when enemy is close. This
is depicted in Figure \ref{fig:4.5}. We will need to find a method to improve the
accuracy of the classifier and make it faster in future works, which is also
mentioned in \ref{sec:5.2}.

\section{Real-time Observations of Adaptive Instance Segmentation}

We will now present observations of highlighted scenarios of adaptive instance
segmentation when using the architecture in real-time.

\subsection{Scenario 1: Leaving out Instances}

\begin{figure}[H]
  \centering
  \includegraphics[width=12cm]{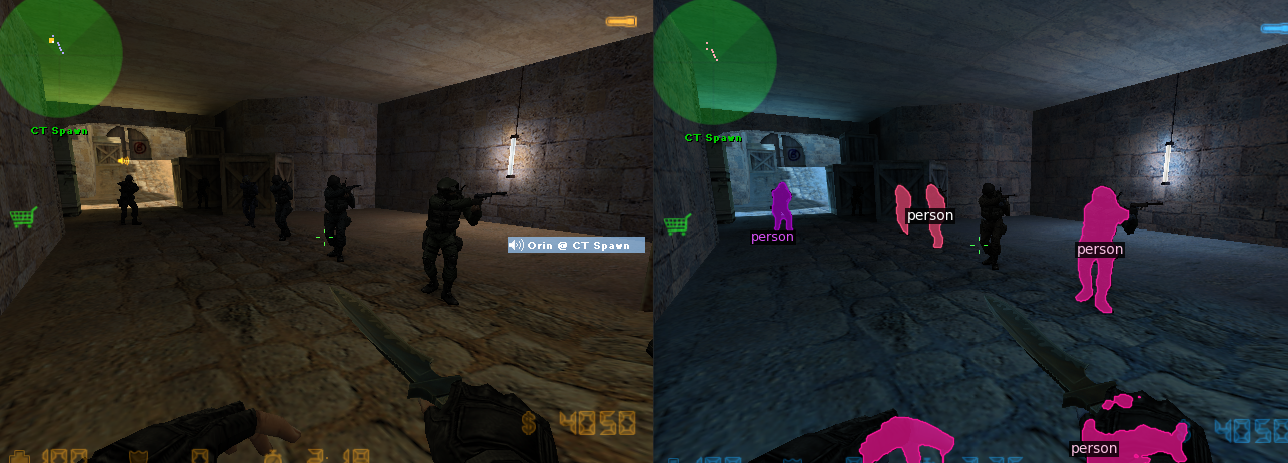}
  \caption{Panoptic segmentation leaves out some instances}
  \label{fig:4.6}
\end{figure}

Sometimes panoptic segmentation leaves out instances when there are more
people in the image. This is shown in Figure \ref{fig:4.6}. The possible cause could
be the shading in the image, making the model difficult to segment correct
instances.

\subsection{Scenario 2: Segmenting our Hand}

\begin{figure}[H]
  \centering
  \includegraphics[width=11cm]{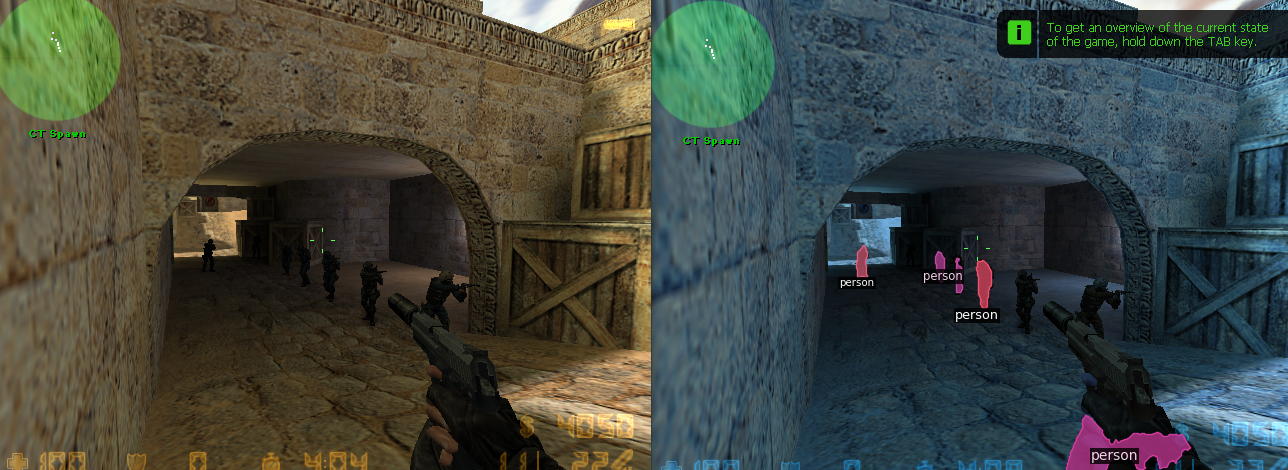}\\[-1pt]
  \includegraphics[width=11cm]{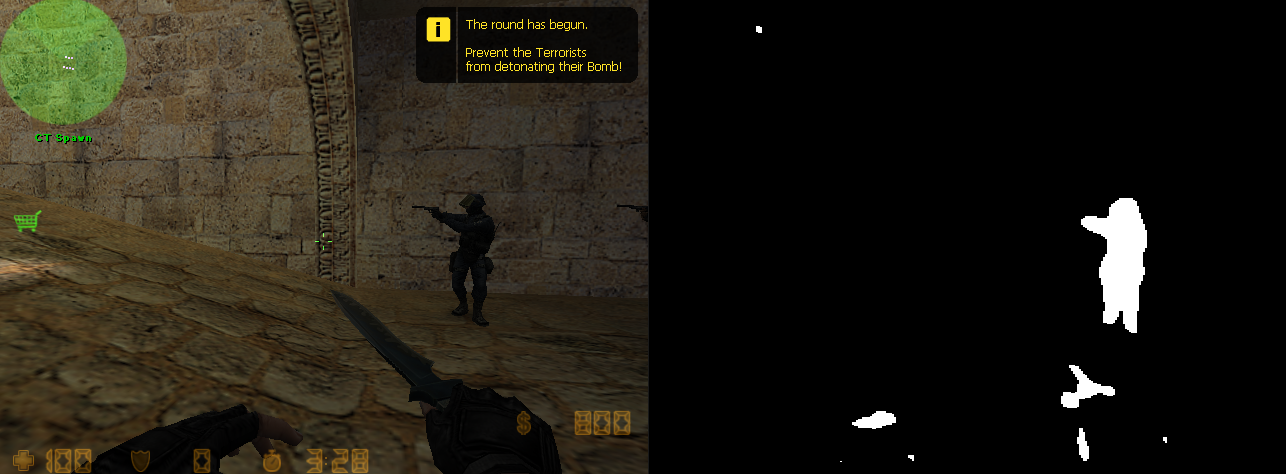}
  \caption{Segmentation algorithms segments our hand. Top: Panoptic segmentation. Bottom: Semantic segmentation.}
  \label{fig:4.7}
\end{figure}

When using Maskformer for panoptic segmentation, sometimes the model
segments our hand as target. Same thing applies to semantic segmentation
but when the player is holding a knife. Figure \ref{fig:4.7} demonstrates this effect.
The reason why this is happening is that Maskformer was trained on detecting
people, therefore it could have confused our hand as a person. For
semantic segmenation, it was caused by the dataset that did not contain
sufficient training data, including player holding a knife.

\subsection{Scenario 3: No Enemies}

\begin{figure}[H]
  \centering
  \includegraphics[width=11cm]{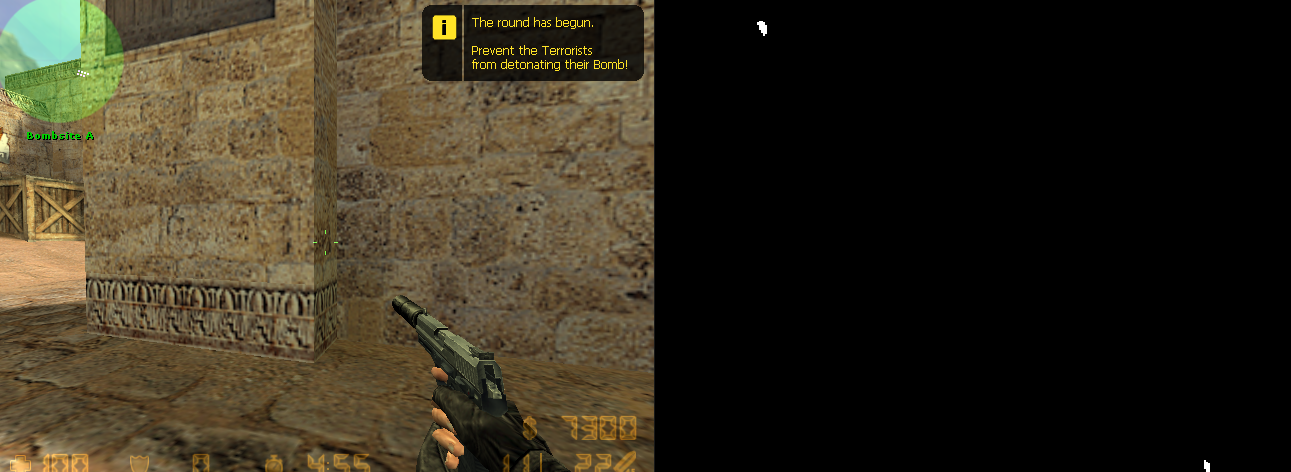}\\[-1pt]
  \includegraphics[width=11cm]{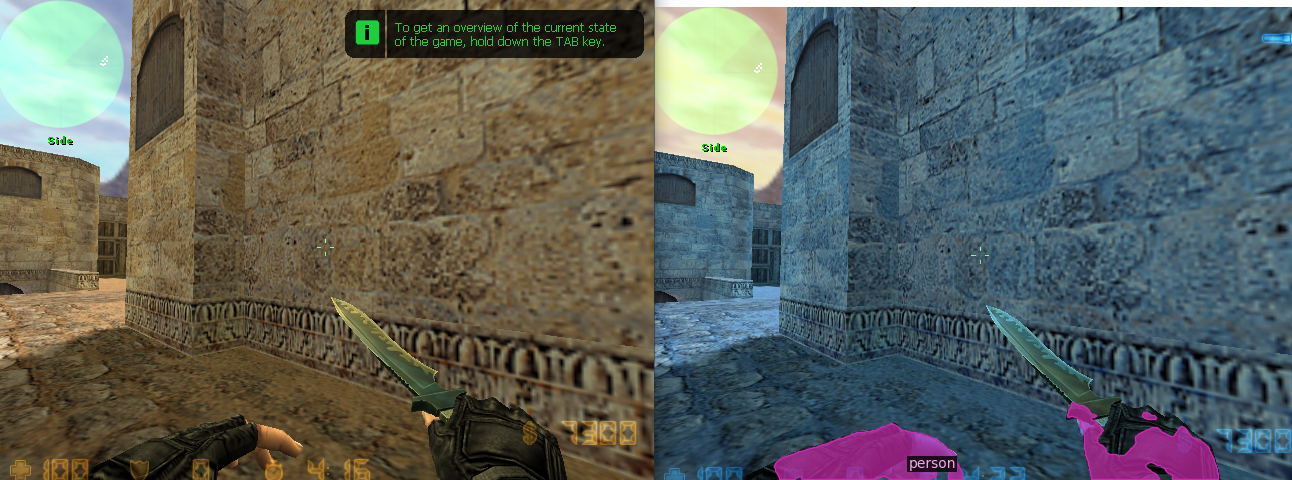}
  \caption{Randomized behaviour when aiming at wall}
  \label{fig:4.8}
\end{figure}

When there are no enemies, for example we are aiming at wall, our model
choose randomly between semantic segmentation and panoptic segmentation,
as shown in Figure \ref{fig:4.8}. If we do not want to have this random behaviour,
we could solve this by adding a new class.

\subsection{Scenario 4: Error in Classifier}

\begin{figure}[H]
  \centering
  \includegraphics[width=11cm]{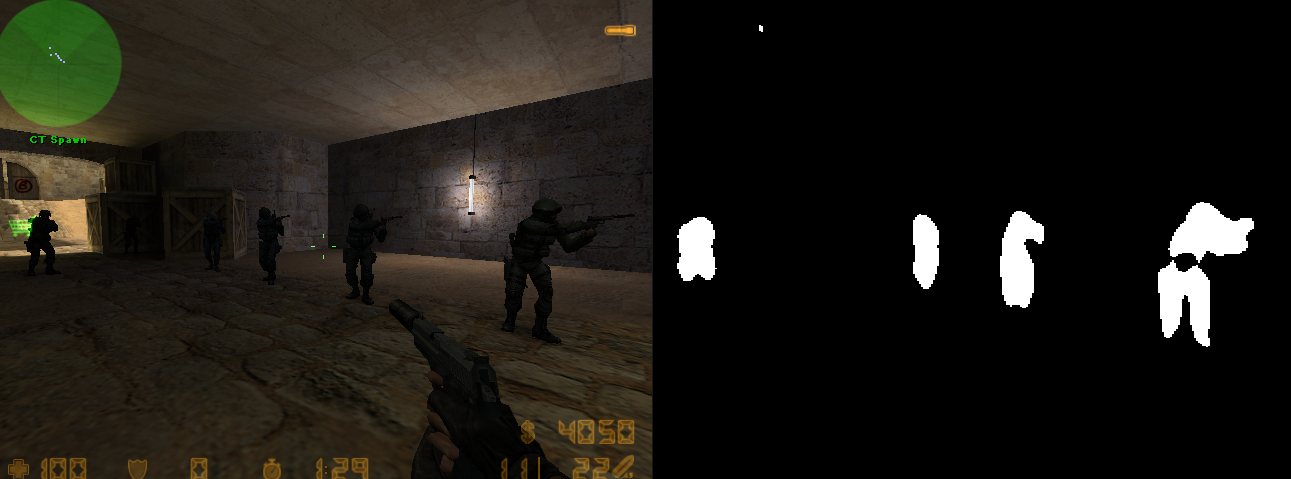}\\[-1pt]
  \includegraphics[width=11cm]{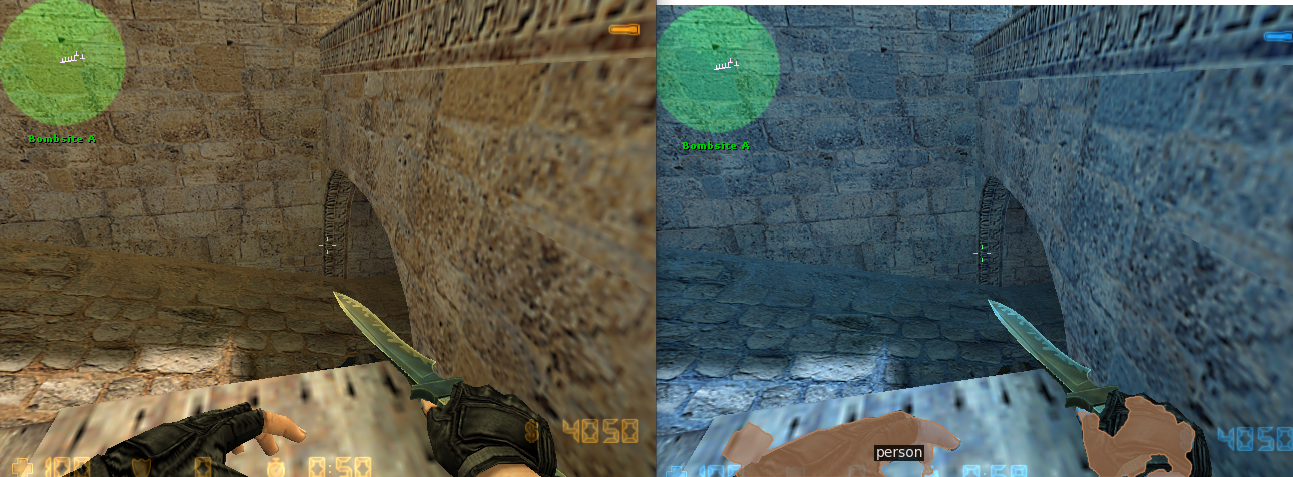}
  \caption{Error in classifier}
  \label{fig:4.9}
\end{figure}

Our classifier will sometimes generate some error, causing it not giving correct
classifications. Figure \ref{fig:4.9} illustrates this effect. We would need a reliable
classifier to remove this error.

\chapter{Discussion}

\section{Dataset}

Our first dataset used to train object detection was \cite{cs16dataset}. It has 248 images
and is split to training, validation, testting sets with 173, 50, 25 images
respectively. Later on, we realize that the size of the dataset had negative
performance impact on object detection since we used a much smaller dataset
in semantic segmenation. For training semantic segmentation, we used small
datasets with 33 images. The dataset contains training, validation, testing
set with 20, 5, 8 images respectively. We note that adding a validation
set was particularly helpful for us to prevent overfitting, using `early stopping'.
Early stopping makes our model stop training when its accuracy is
degrading. This means that the model is in the global minimum, therefore
preventing overfitting. We then used similar dataset to train object detection
and conclude that object detection is faster than semantic segmentation.
For classifier, we used a larger datasets for both adaptive distance detection
and adaptive instance segmentation. However, adding a validation set was
not enough to prevent overfitting. We decided to add additional augmentations
with rotation, horizontal flipping and brightness change to generalize
the model. For testing adaptive distance detection and adaptive instance
segmentation we created two additional testing datasets. Both datasets use
the annotations of semantic segmentation to compare the accuracy of algorithms.
Dataset for adaptive distance detection contains 10 images with half
of the images having the enemy far away, and other half near. Same concept
is applied to testing dataset for adaptive instance segmentation, having 8
images, with half of the images with more enemies, other half one enemy.
Images can be downloaded, as shown Appendix \ref{app:A}.

\section{Limitations and Future Research}
\label{sec:5.2}

New architecture have been proposed in this paper. We successfully created
adaptive distance detection and adaptive instance segmentation. However,
there exists number of limitations in the architecture.

The idea of adaptive instance segmentation was proposed but not verified
to be beneficial in real use. This is because, more experimentations are
needed to verify if there is going to be an exponential blowup of inference
time on panoptic method compared to semantic method when model is more
complex. This encourages to compare the inference time of equally complex
models of both algorithms.

Both classifiers for adaptive algorithms could be improved, as we often
see the inaccuracy of model predicting of the correct output. Both classifiers
used larger datasets for training but had errors, as shown in Evaluation
chapter. This encourages us to find methods to create a reliable fast classifier
giving a limited small dataset. Attaching a classifier can be generalized
to be trained on more than one game. This generalization can let us use
classifier on new games without specifically training on it. However, we need
to carefully choose the method of training classifier.

For panoptic segmentation, although we observed that Maskformer was
accurate at segmenting enemies, it does have some drawbacks. Firstly, at
scenarios where there is shading, it is possible for the model to leave some
instances. Secondly, it sometimes segments our hand as an enemy because
the model was pretrained to detect people instead of enemy. For future
research, we plan to create a custom panoptic model that can detect enemies
without segmenting our hand.

The dataset for object detection and semantic segmentation could be
scaled up, in which it will boost the accuracy. Because we used a small
dataset to train both segmentation and object detection models, both models
are less accurate. We presented scenarios, in which models had some pitfalls
during observations. Nevertheless, most of these problems can be solved
by scaling up data, making the dataset more diverse. Additionally, it is
desirable to see the change of inference time if the dataset is scaled up.
Finally, We plan to use the information from adaptive distance detection
and adaptive instance segmentation to create a bot that is able to shoot at
enemies in the near future.

\chapter{Conclusion}

Our project firstly compared the inference time of semantic segmentation
and object detection. We concluded that the inference time of YOLOv5s is
better than the custom semantic segmentation. We found that there is a
problem of exponential blowup of inference time of segmentation algorithm
compared to object detection. This paper showed the solution of combining
two detection algorithms by adding a bridge of classifier to benefit from
latency performance of YOLOv5s and accuracy of semantic segmentation
since semantic segmentation detects object pixel by pixel. Object detection
algorithm was trained on our custom dataset using YOLOv5s and semantic
segmentation was trained using U-Net architecture. A custom classifier
was then added to these algorithms. The added classifier was to classify the
distance of an object. Additionally, the same concept was applied to semantic
segmentation and panoptic segmentation. This time, the classifier was
trained to use semantic segmentation when there is one instance of enemy
and panoptic segmentation when there are more instaces of enemies.


\appendix

\chapter{How to run our code}
\label{app:A}

\begin{figure}[H]
  \begin{minipage}[c]{0.48\textwidth}
    \centering
    \includegraphics[width=6cm]{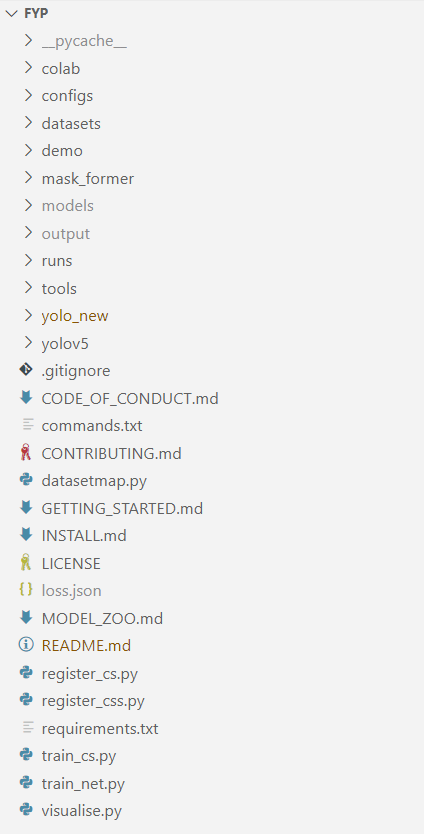}
    \caption{Structure of our project}
    \label{fig:A.1}
  \end{minipage}\hfill
  \begin{minipage}[c]{0.48\textwidth}
    \centering
    \includegraphics[width=6cm]{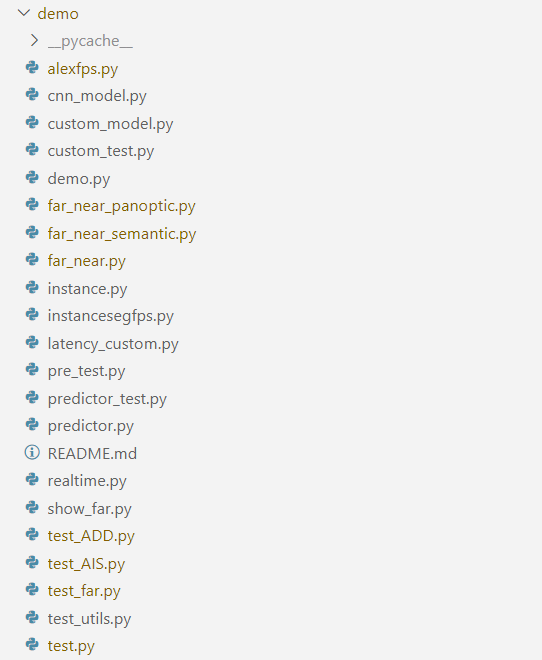}
    \caption{Files in demo folder}
    \label{fig:A.2}
  \end{minipage}

  \vspace{1em}
  \begin{minipage}[c]{0.48\textwidth}
    \centering
    \includegraphics[width=5cm]{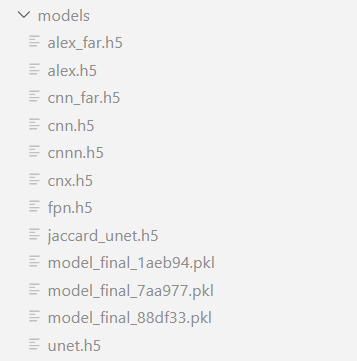}
    \caption{Models}
    \label{fig:A.3}
  \end{minipage}\hfill
  \begin{minipage}[c]{0.48\textwidth}
    \centering
    \includegraphics[width=8cm]{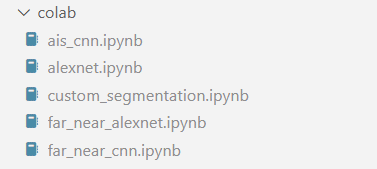}
    \caption{Colab notebooks}
    \label{fig:A.4}
  \end{minipage}
\end{figure}

The structure of our project repository will look like Figure \ref{fig:A.1}. Our code
is in the folder demo and has following files shown in Figure \ref{fig:A.2}. Our
main focus would be on far\_near\_semantic.py (implementation for ADD)
and instancesegfps.py (implementation for AIS). These files can be run by
commands in Listings \ref{lst:A.1} and \ref{lst:A.2} at the root folder.

\begin{lstlisting}[style=shell, caption={bash version}, label={lst:A.1}]
python .\demo\instancesegfps.py --config-file
configs\ade20k-150-panoptic\
  maskformer_panoptic_R50_bs16_720k.yaml
--opts MODEL.WEIGHTS models\model_final_7aa977.pkl
\end{lstlisting}

\begin{lstlisting}[style=shell, caption={bash version}, label={lst:A.2}]
python .\demo\far_near_semantic.py
\end{lstlisting}

Note that these files will use classes defined in cnn\_model.py and custom\_model.py
which will need models stored in \texttt{\textbackslash models}. These models can
be downloaded using the link provided in README.md. Figures \ref{fig:A.3} shows
the list of models that could be downloaded. Make sure you have Counter
Strike: Condition Zero displayed in a window with a resolution of 640x480.
The window should be placed in the top-left corner of your screen.

The colab notebooks in which we trained our classifier and segmentation
models are stored in \texttt{\textbackslash colab}, as shown in Figure \ref{fig:A.4}.

\begin{figure}[H]
  \centering
  \includegraphics[width=15cm]{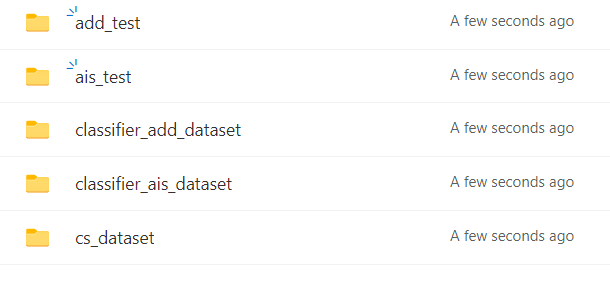}
  \caption{List of datasets}
  \label{fig:A.5}
\end{figure}

Finally, the datasets can be download from links in README.md. Figure
\ref{fig:A.5} shows the list of datasets.

\chapter{YOLOv5s with 2 classes on a small dataset}
\label{app:B}

\begin{figure}[H]
  \centering
  \includegraphics[width=13cm]{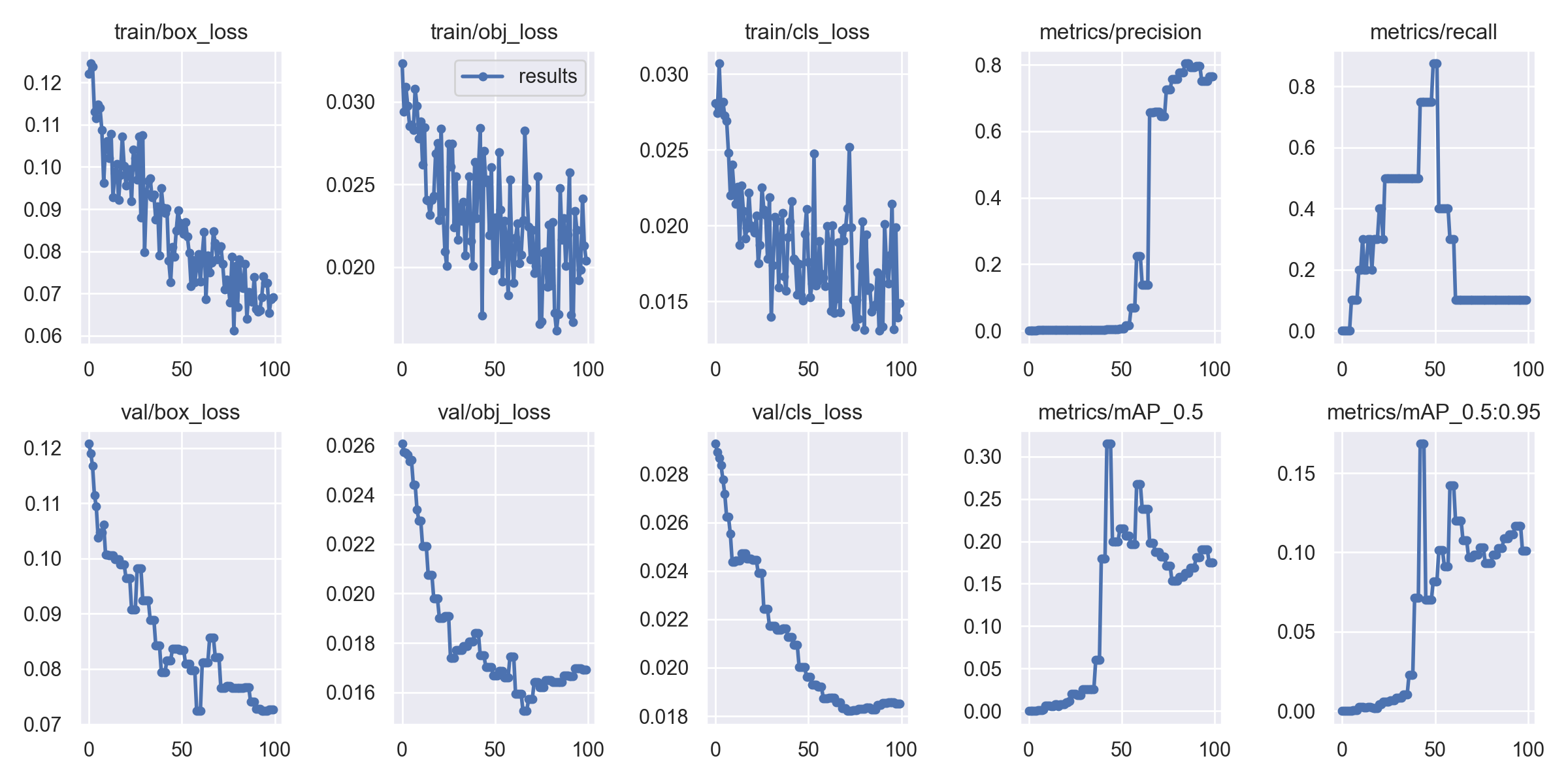}
  \caption{Loss and accuracy}
  \label{fig:B.1}
\end{figure}

\begin{figure}[H]
  \centering
  \includegraphics[width=10cm]{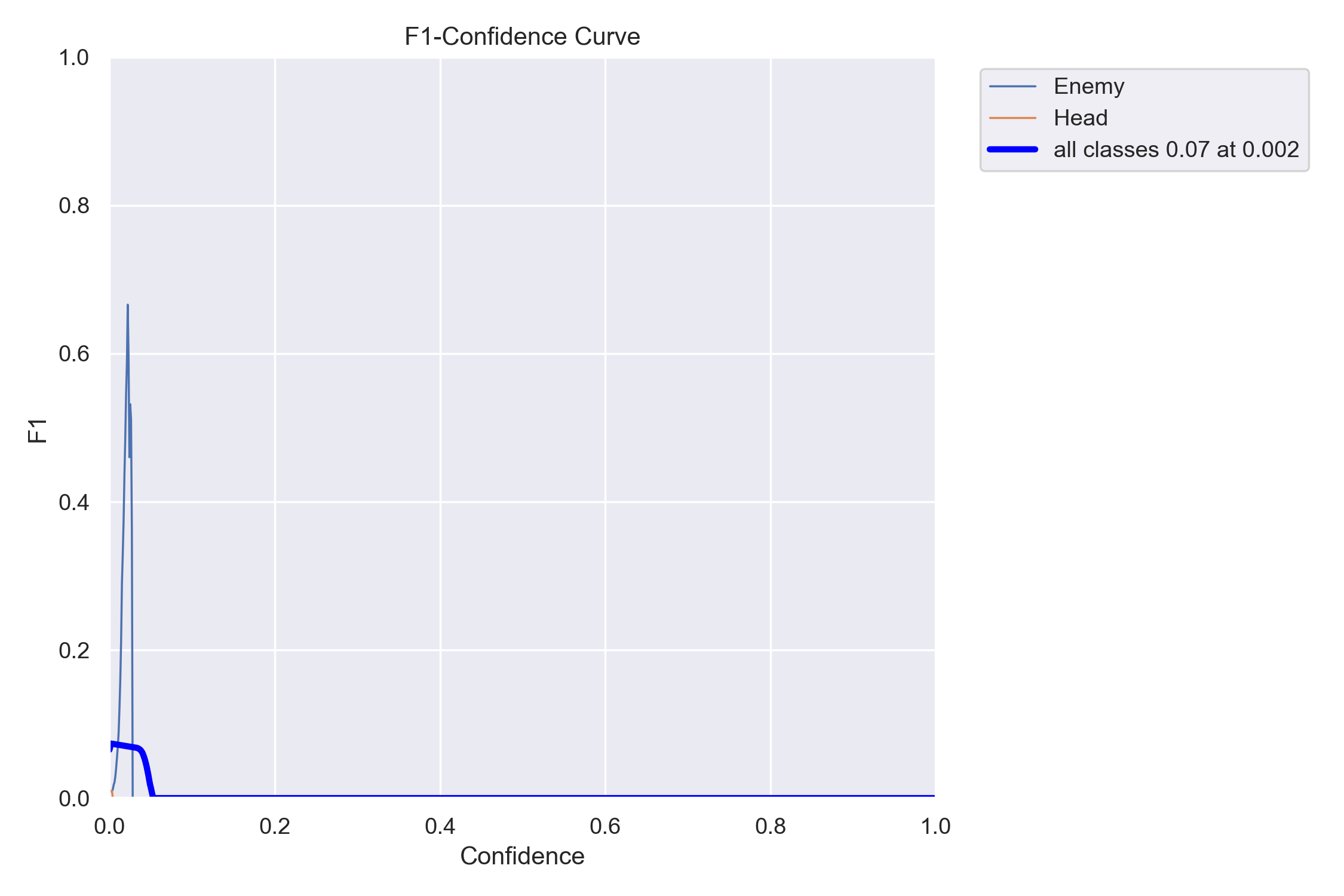}
  \caption{F1 score}
  \label{fig:B.3}
\end{figure}

\begin{figure}[H]
  \centering
  \includegraphics[width=10cm]{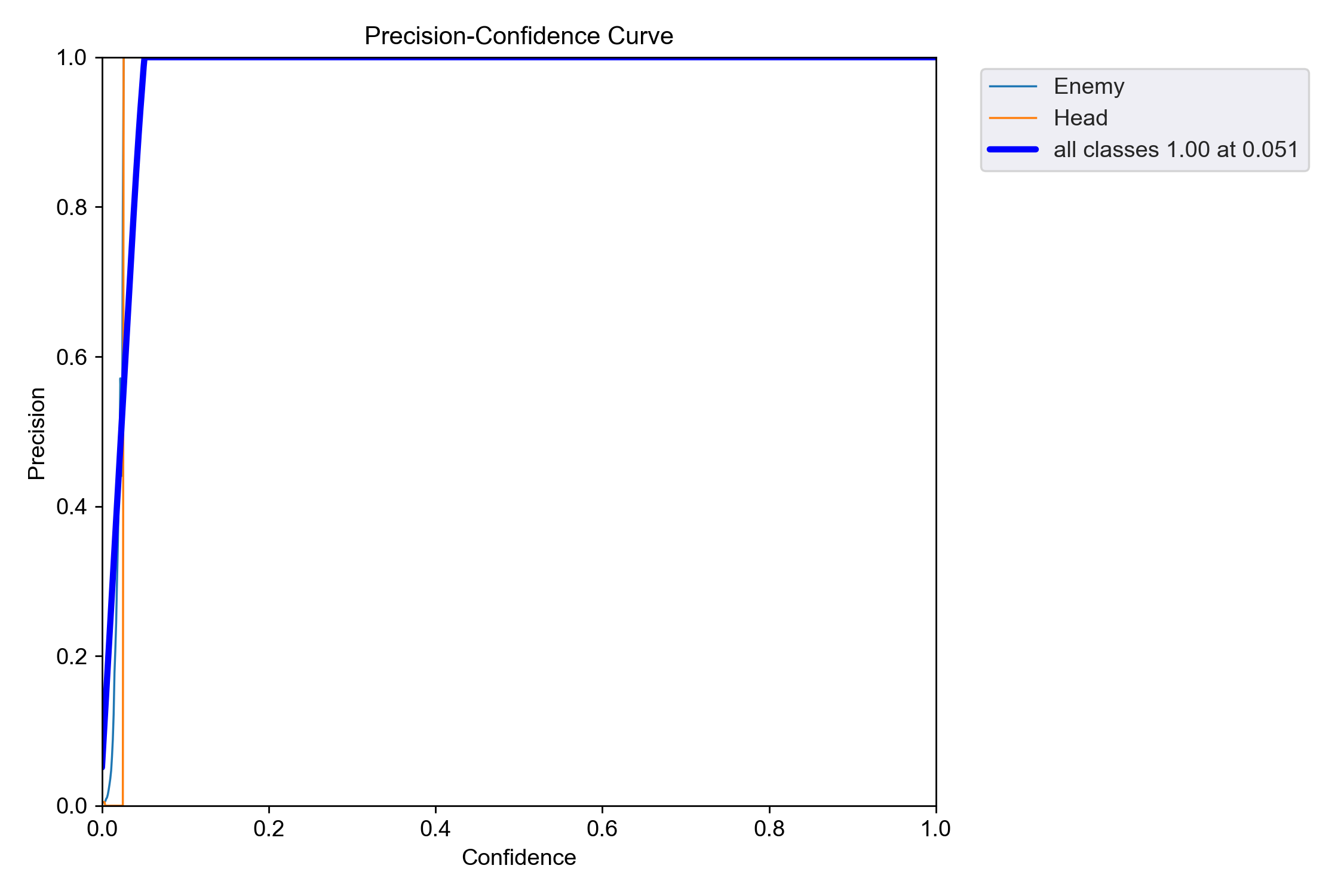}
  \caption{Precision to confidence}
  \label{fig:B.4}
\end{figure}

\begin{figure}[H]
  \centering
  \includegraphics[width=10cm]{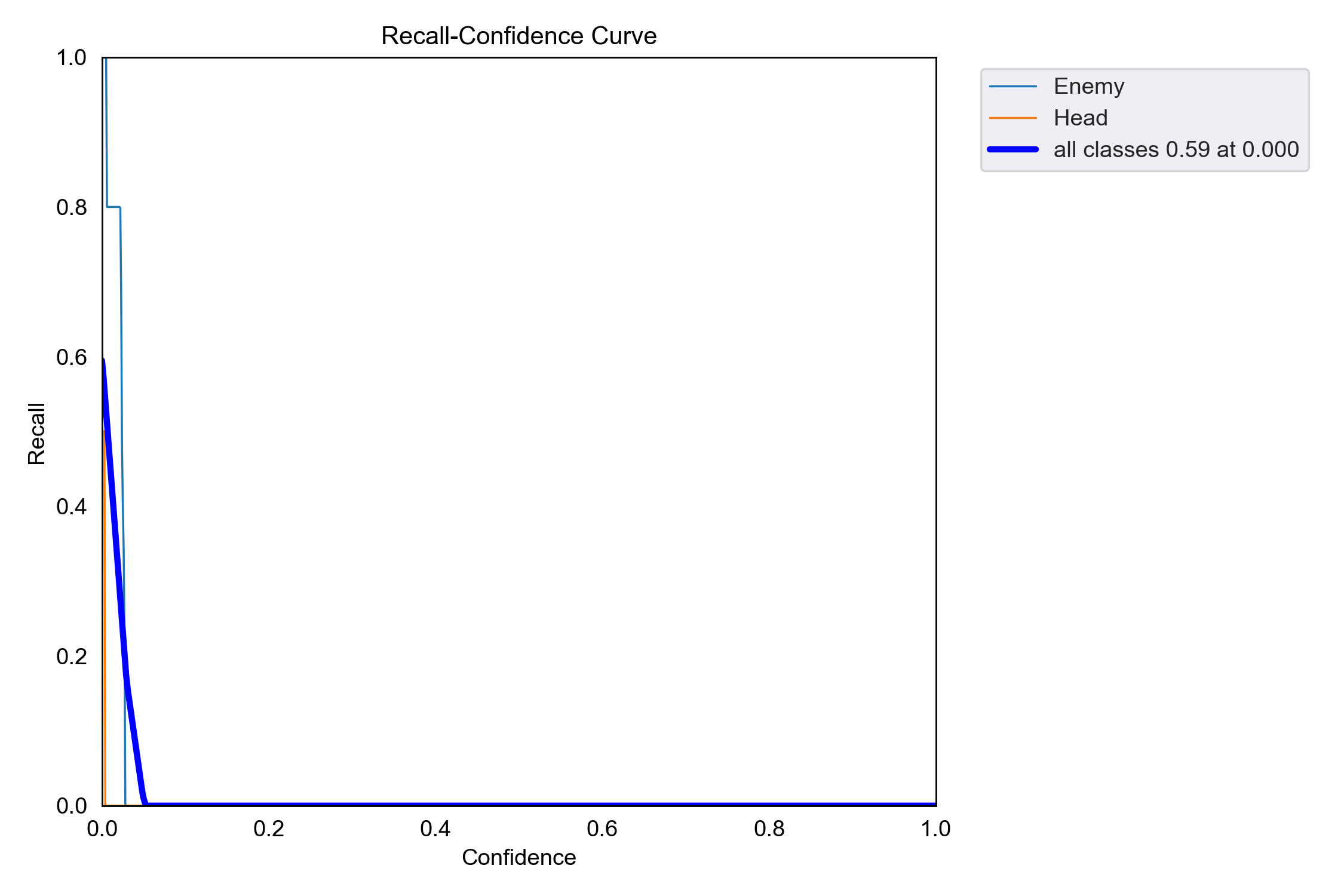}
  \caption{Recall to confidence}
  \label{fig:B.5}
\end{figure}

We trained our object detection with a small dataset but with 2 classes.
Figure \ref{fig:B.1} demonstrates the loss of our model. We observe that cls\_loss is
not constant anymore since we trained with 2 classes. We notice that the
peak of F1-curve is at 0.002, as depicted in Figure \ref{fig:B.3}. Figures \ref{fig:B.4} and \ref{fig:B.5}
are precision to connfidence and recall to confidence. Precision achieved 1.00
accuracy at 0.385 confidence, whereas recall achieved 0.57 accuracy at 0.00
connfidence. It is evident that it was harder to predict with 2 classes.

\begin{figure}[H]
  \centering
  \includegraphics[width=10cm]{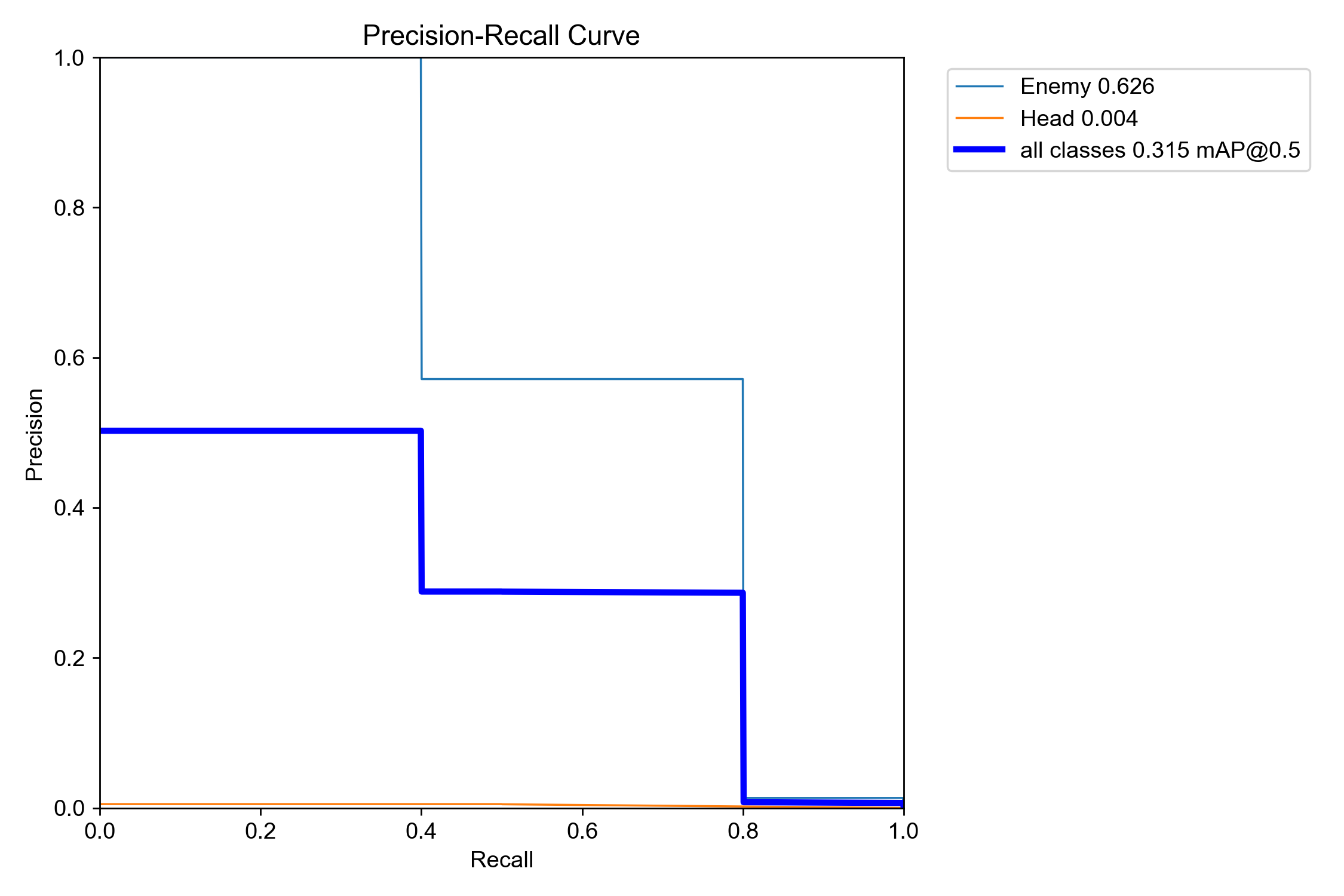}
  \caption{Precision to recall}
  \label{fig:B.2}
\end{figure}

\end{document}